\documentclass[journal,twocolumn]{IEEEtran}

\usepackage{times}
\usepackage{epsfig}
\usepackage{graphicx}
\usepackage{amsmath}
\usepackage{amssymb}
\usepackage{algorithmic}
\usepackage[linesnumbered,ruled,vlined]{algorithm2e}
\usepackage{multirow}
\usepackage{booktabs}
\usepackage{siunitx}
\usepackage{bm}
\usepackage{amsfonts}
\usepackage{caption} 
\usepackage{dsfont}
\usepackage{url}
\usepackage[margin=1.41cm]{geometry}

\usepackage{CJK}
\newif\ifdraft\drafttrue
\ifdraft

\newcommand\liming[1]{{\huge \color{black}#1  \textbf{}}}
\fi
\usepackage[pagebackref=true,breaklinks=true,letterpaper=true,colorlinks,bookmarks=false]{hyperref}

\ifCLASSINFOpdf

\else

\fi

\begin{document}
\begin{CJK*}{UTF8}{gkai}

\title{ \liming{Beyond Batch Learning: Global Awareness Enhanced Domain Adaptation }}

\author{Lingkun Luo, Shiqiang Hu, ~Liming~Chen,~\IEEEmembership{Senior~member,~IEEE}      

\thanks{K. Luo and S. Qiang are from the School of Aeronautics and Astronautics, Shanghai Jiao Tong University, 800 Dongchuan Road, Shanghai, China e-mail: (lolinkun1988,sqhu)@sjtu.edu.cn; L. Chen from LIRIS, CNRS UMR 5205, Ecole Centrale de Lyon, and Institut Universitaire de France (IUF), 36 avenue Guy de Collongue, Ecully,  France e-mail: liming.chen@ec-lyon.fr. \textbf{\textcolor{blue}{This paper has been accepted for publication in IEEE TPAMI in 2025.}}}
}

\markboth{Journal of \LaTeX\ 2021}%
{Shell \MakeLowercase{\textit{et al.}}: Bare Demo of IEEEtran.cls for IEEE Journals}

\maketitle

\begin{abstract} In domain adaptation (DA), the effectiveness of deep learning-based models is often constrained by batch learning strategies that fail to fully apprehend the global statistical and geometric characteristics of data distributions. Addressing this gap, we introduce "Global Awareness Enhanced Domain Adaptation" (GAN-DA), a novel approach that transcends traditional batch-based limitations. GAN-DA integrates a unique predefined feature representation (PFR) to facilitate the alignment of cross-domain distributions, thereby achieving a comprehensive global statistical awareness. This representation is innovatively expanded to encompass orthogonal and common feature aspects, which enhances the unification of global manifold structures and refines decision boundaries for more effective DA. Our extensive experiments, encompassing 27 diverse cross-domain image classification tasks, demonstrate GAN-DA's remarkable superiority, outperforming 24 established DA methods by a significant margin. Furthermore, our in-depth analyses shed light on the decision-making processes, revealing insights into the adaptability and efficiency of GAN-DA. This approach not only addresses the limitations of existing DA methodologies but also sets a new benchmark in the realm of domain adaptation, offering broad implications for future research and applications in this field.

\end{abstract}


\IEEEpeerreviewmaketitle

	\section{Introduction}
	\label{Introduction}

\IEEEPARstart{T}{he} success of deep learning models largely hinges on the availability of abundant, well-labeled training data (denoted as ${{\mathcal{X}_\mathcal{S}}},{{\mathcal{Y}_\mathcal{S}}}$), which drives their impressive performance on test samples ($\mathcal{X}_\mathcal{T}$). These models typically operate under the assumption that training and testing data are independent and identically distributed (iid). This iid assumption implies that the data in both training and testing sets are drawn from the same statistical distribution.

However, real-world scenarios often challenge this assumption \cite{pan2010survey, DBLP:journals/csur/LuLHWC20}. For instance, variations in environmental conditions (like changes in lighting or temperature) or differences in data capture devices can lead to significant discrepancies between training (source domain) and testing data (target domain). These discrepancies create a divergence from the iid assumption, posing a critical challenge: ensuring that a model trained on a specific dataset (the source domain) maintains its effectiveness when applied to a different dataset (the target domain).

Addressing this challenge is the primary objective of domain adaptation research. This field focuses on developing methodologies that enable models to transfer knowledge from one domain to another, effectively bridging the gap created by non-iid data. Our work contributes to this field by proposing a novel approach that enhances domain adaptation through global awareness, optimizing models to perform robustly across diverse data distributions.

In the domain adaptation landscape, there are two main research paradigms: semi-supervised domain adaptation, assuming a certain amount of labeled data is available in the target domain, and unsupervised domain adaptation (UDA), where the target domain has no labels. Our paper specifically focuses on UDA. State-of-the-art UDA methods fall into two categories: shallow domain adaptation and deep domain adaptation (DL-DA) methods.

The conventional \textit{shallow} \textbf{DA} approaches majorly rely on analytical optimization \cite{wang2018visual, long2013adaptation} or the Rayleigh quotient optimization \cite{long2013transfer, luo2020discriminative}, both of which require global optimization of the entire cross-domain samples, thereby face challenges with scalability. For example, the newly designed challenging \textbf{DA} dataset, \textit{i.e.} \textbf{DomainNet} \cite{peng2019moment}, presents computational challenges for shallow \textbf{DA} approaches. Specifically, in the \emph{Quickdraw $\rightarrow$ Real} cross-domain task, using the entire \textbf{DomainNet} dataset requires computations on a $(3.4*{10^6})*(3.4*{10^6})$ size matrix, which is infeasible due to the high computational complexity and hardware requirements. As a result, these \textbf{DA} approaches,  sensitive to the dataset size, face the issue of scaling,  hindering further developments of this research paradigm for large cross-domain datasets. On the other hand, DL-DA techniques, such as \cite{long2015learning, ganin2016domain, tzeng2017adversarial}, have addressed scalability concerns through batch learning, achieving impressive domain adaptation performance. However, in reducing complexity through iterative optimization over small batches of data, batch learning also introduces a global ignorance problem in terms of holistic statistical and geometric representations, hindering effective optimization of certain critical terms in the domain adaptation error bound.

To illustrate this issue, we review a cornerstone theoretical result in domain adaptation \cite{ben2010theory}, which provides an error-bound estimation of a learned hypothesis $h$ on a target domain : 

\begin{equation}\label{eq:bound}
		\resizebox{0.9\hsize}{!}{
			$\begin{array}{l}
			{e_{\cal T}}(h) \le {e_{\cal S}}(h) + {d_{\cal H}}({{\cal D}_{\cal S}},{{\cal D}_{\cal T}})+ \\
			\;\;\;\;\;\;\;\; \min \left\{ {{{\cal E}_{{{\cal D}_{\cal S}}}}\left[ {\left| {{f_{\cal S}}({\bf{x}}) - {f_{\cal T}}({\bf{x}})} \right|} \right],{{\cal E}_{{{\cal D}_{\cal T}}}}\left[ {\left| {{f_{\cal S}}({\bf{x}}) - {f_{\cal T}}({\bf{x}})} \right|} \right]} \right\}
			\end{array}$}
	\end{equation}

Eq.(\ref{eq:bound}) states that the performance ${e_{\cal T}}(h)$ of a hypothesis $h$ on the target domain is bounded by the following three terms: \textbf{Term.1} ${e_{\cal S}}(h)$ which denotes the source domain structural risk; \textbf{Term.2} ${{d_{\cal H}}({{\cal D}_{\cal S}},{{\cal D}_{\cal T}})}$ which measures the $\mathcal{H}$\emph{-divergence}\cite{kifer2004detecting} between two distributions ($\mathcal{D_S}$, $\mathcal{D_T}$); \textbf{Term.3} is the last term which characterizes the difference in labeling functions across the two domains. Within the \textbf{DL-UDA} paradigm, while minimization of \textbf{Term.1} can benefit from the well-labeled source domain which offers high-quality functional regularization,  effective optimization of  \textbf{Term.2} and \textbf{Term.3} requires holistic knowledge of data, thereby poses a serious challenge for batch learning-driven deep \textbf{DA} methods due to the lack of global awareness regarding statistical and geometric representations.

\begin{figure}[h!]

	\centering
	\includegraphics[width=1\linewidth]{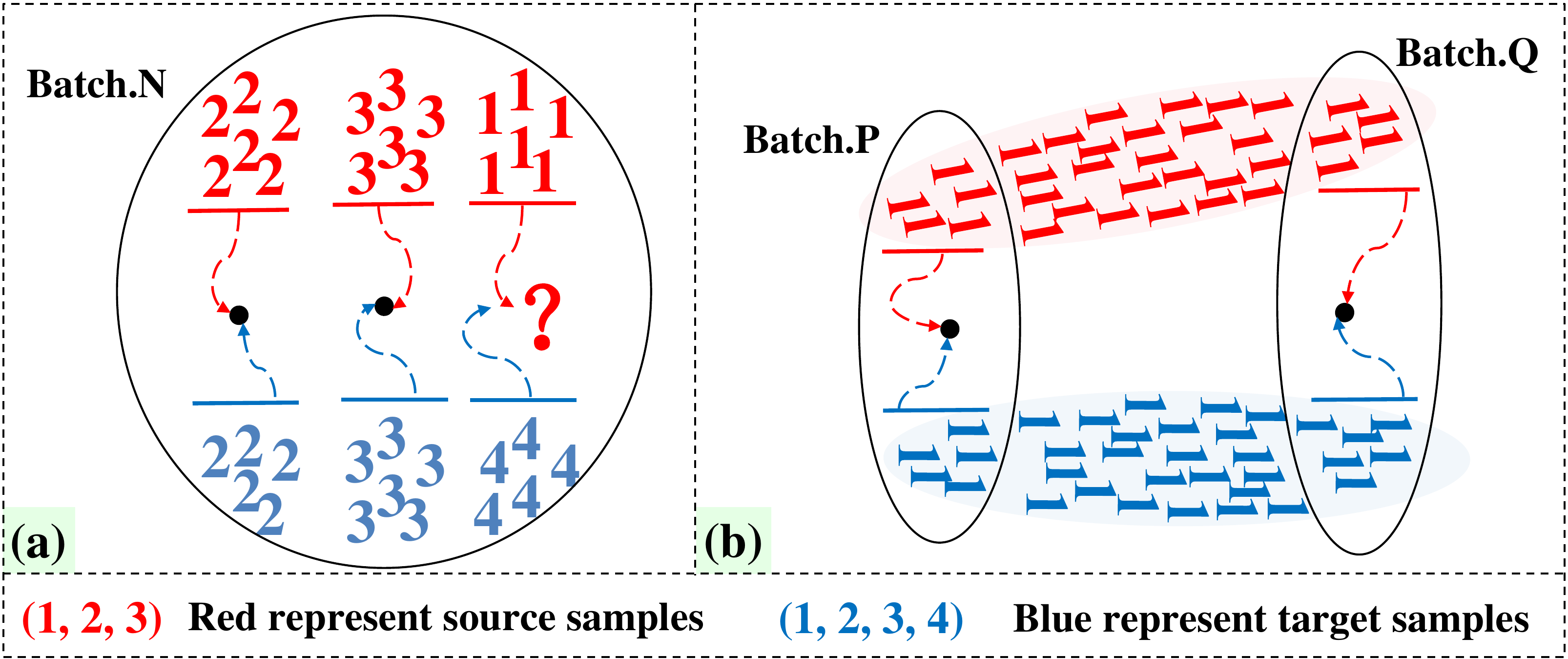}
	\caption{ Batch learning-induced unreliable functional learning within (Fig.\ref{fig:1}.(a)) or without (Fig.\ref{fig:1}.(b)) a single batch.} 
	\label{fig:1}
\end{figure}

\begin{figure*}[h!]
	\centering
	\includegraphics[width=1\linewidth]{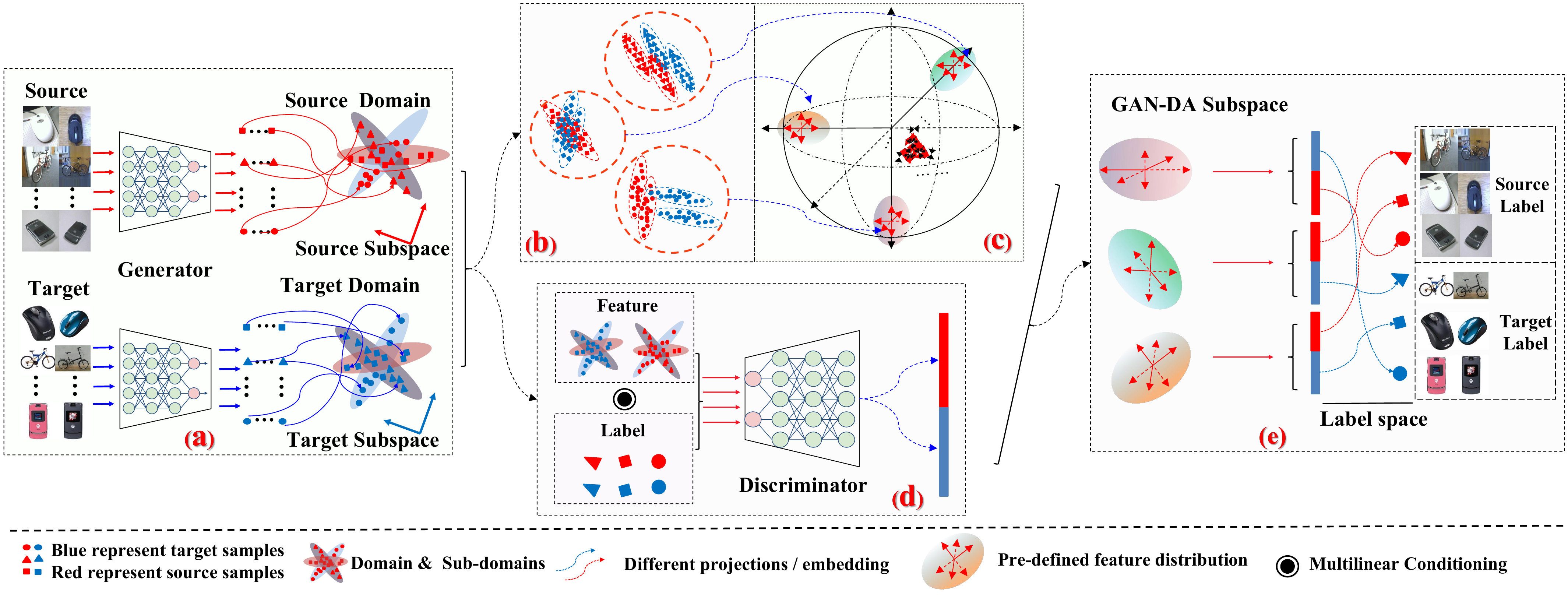}
	\caption {Illustration of the proposed \textbf{GAN-DA} method. In Fig.\ref{fig:3} (a), we show examples of source (in red) and target (in blue) data such as mouse, bike, and smartphone images, which have different distributions and inherent hidden data geometric structures. Different geometric shapes represent samples of different class labels. Fig.\ref{fig:3} (b\&c) depicts the aligned data distributions by using the generator and the predefined feature representations. Fig.\ref{fig:3} (d) illustrates harmoniously blending of the feature/label space using conditional adversarial learning strategies. Fig.\ref{fig:3} (e) shows the final classification results.}
	\label{fig:3}
\end{figure*}

Specifically, we illustrate in Fig.\ref{fig:1} the limitations of batch learning in \textbf{DL-DA} arising from the disregard of global data statistical and geometric properties.

	\begin{itemize}

     \item  {As shown in Fig.\ref{fig:1}, red characters represent samples from the source domain, while blue characters represent samples from the target domain. Specifically, the red characters '\textcolor{red}{\textbf{1,2,3}}' and the blue characters '\textcolor{blue}{\textbf{1,2,3,4}}' denote samples from distinct cross-subdomains. In Fig.\ref{fig:1}.(a), the red and blue samples within the black circle are drawn from a single batch, \textbf{Batch.N}. In contrast, Fig.\ref{fig:1}.(b) shows red and blue samples within two separate black circles, which are drawn from two distinct batches, \textbf{Batch.P} and \textbf{Batch.Q}, respectively.}

    \item  {As illustrated in Fig.\ref{fig:1}.(a), due to incomplete data sampling in \textbf{Batch.N}, there is typically insufficient data to ensure symmetrical cross-subdomain alignment for cross-domain conditional distribution alignment. Consequently, neglecting global statistical awareness and forcibly aligning the source subdomain '\textcolor{red}{\textbf{1}}' with the target subdomain '\textcolor{blue}{\textbf{4}}' can lead to negative transfer.}

    \item   {Figure \ref{fig:1}.(b) shows two batches of data where both source and target samples are from subdomain \textbf{1}. Due to the limitations of batch learning in capturing global cross-domain statistics and geometric knowledge, the learned embeddings (represented as black dots) may appear in different locations within the optimized common feature space, which can hinder the convergence of the cross-domain decision boundary optimization.}

    \end{itemize}

To deal with the aforementioned issue of global ignorance in \textbf{DL-DA}, we introduce in this paper a novel deep \textbf{DA} approach, namely \textbf{G}lobal \textbf{A}wareness e\textbf{N}hanced deep \textbf{D}omain \textbf{A}daptation (\textbf{GAN-DA}), as visualized in Fig.\ref{fig:3}. Specifically, we introduce in \textbf{GAN-DA} a predefined feature representation (\textbf{PFR}) (Fig.\ref{fig:3}.(c)) which empowers \textbf{DA} models with global \textit{statistic awareness}, irrespective of the \textit{batch-learning} mechanism. Furthermore, we extend \textbf{PFR} to encompass \textbf{orthogonal feature representation} and \textbf{common feature representation} terms, proactively enhancing global \textit{geometric awareness}. As a result of these predefined data representation structures, \textbf{GAN-DA} facilitates a comprehensive comprehension of global data representations. Our contributions can be summarized as follows:

\begin{itemize}
	\item  We evidence the issue of global ignorance in \textbf{DL-DA} methods when performing batch learning-baed optimization. 
 
    \item  We propose a novel \textbf{DL-DA} method, namely \textbf{GAN-DA} with 1): a predefined feature representation (\textbf{PFR}) term to imbue the model with global \textit{statistic awareness}, thereby  effectively aligning cross-domain distributions and optimizing \textbf{Term.2} in Eq.(\ref{eq:bound}); and 2) extension of \textbf{PFR} to encompass \textit{orthogonal feature representation} and \textit{common feature representation} terms which promote the unification of global manifold structure and facilitate decision boundary optimization, thereby reducing \textbf{Term.3} in Eq.(\ref{eq:bound}).

	\item We conducted extensive experiments on 27 image classification domain adaptation tasks across 4 benchmarks, demonstrating the superior performance of our proposed \textbf{GAN-DA} method compared to 24 state-of-the-art algorithms. We further validated our model using synthetic and real datasets, conducting an ablation study for deeper insights. By analyzing recognition accuracy and feature visualization results from our derived partial models, we gained a better understanding of the effectiveness of our proposed \textbf{DA} model.

\end{itemize}

The paper is organized as follows. Sect.\ref{Related Work} discusses the related work. Sect.\ref{The proposed method} presents the method. Sect.\ref{Experiments} benchmarks the proposed \textbf{DA} method and provides an in-depth analysis. Sect.\ref{Conclusion} draws the conclusion.

\section{Related Work}
\label{Related Work}

Cutting-edge \textbf{DA} techniques encompass two primary research avenues: 1) Shallow \textbf{DA}; and 2) Deep \textbf{DA}. In Section \ref{subsect: Shallow DA} and Section \ref{subsect: Deep DA}, we provide an insightful overview of these techniques and simultaneously engage in a comparative discussion with our innovative \textbf{GAN-DA} approach. This parallel exploration aims to better clarify the underlying rationale of our proposed method.

\subsection{Shallow Domain Adaptation}
\label{subsect: Shallow DA}

Given the error bound of the target domain, we observed that \textbf{Term.1} in Eq.(\ref{eq:bound}) is generally been optimized using a supervised learning approach. Consequently, shallow \textbf{DA} methods primarily focus on reducing the domain shift by employing statistic alignment or geometric regularizations to explicitly minimize \textbf{Term.2} and \textbf{Term.3} in Eq.(\ref{eq:bound}).

\subsubsection{Statistic Alignment-based DA \textbf{(STA-DA)}} 
\label{Statistic alignment-based DA}

The core principle of \textbf{STA-DA} rests on the premise that seeking an optimized common feature space through statistical measurements can significantly mitigate cross-domain disparities. As a result, the knowledge acquired from the source domain can be seamlessly applied to the target domain within this fine-tuned common feature space.

To this end, recent research has incorporated a range of statistical measurements, \textit{e.g.}, Bregman Divergence \cite{4967588}, Wasserstein distance \cite{courty2017joint, courty2017optimal}, and \textit{Maximum Mean Discrepancy} (\textbf{MMD}) measurement \cite{pan2011domain}, to identify a common feature space that reduces cross-domain divergence, thereby diminishing Term.2 in Equation (\ref{eq:bound}). Specifically, through the application of dimensionality reduction techniques, \textbf{TCA} \cite{pan2011domain} explicitly addresses the disparity between source and target domains in terms of marginal distribution \cite{4967588} \cite{pan2011domain}. \textbf{JDA} \cite{long2013transfer} builds upon the framework of \textbf{TCA} and further leverages conditional distribution alignment for effective sub-domain adaptation. In contrast to prior \textbf{STA-DA} methods, \textbf{ILS} \cite{herath2017learning} focuses on learning a discriminative latent space using the Mahalanobis metric to align statistical properties across different domains. Beyond the previous divergence reduction-based \textbf{DA} approaches, recent research argues in favor of '\textit{preserving divergence within differently labeled sub-domains to simplify the next round of classifier optimization}'. To address this, \cite{lu2018embarrassingly} explores the discriminative contribution by drawing on the advantages of linear discriminant analysis (\textbf{LDA}) to optimize the common feature space. \textbf{DGA-DA} \cite{luo2020discriminative} underscores the effectiveness of discrimination, introducing \textit{‘repulsive force’} terms as an additional element. \textbf{ARG-DA} \cite{luo2022attention} aligns cross sub-domains in a discriminative manner using a specifically designed attention-regulated graph.

While \textbf{STA-DA} techniques utilize statistical measurements to mitigate domain shifts across distinct kernel spaces, a notable limitation lies in their failure to provide a geometric perspective for addressing \textbf{Term.3} of Equation (\ref{eq:bound}). Moreover, the growing volume of cross-domain samples results in a substantial reduction in computational efficiency for the optimization of the overall kernel space, which hampers the further advancement of \textbf{STA-DA} methods within the shallow \textbf{DA} paradigm.

\subsubsection{Geometric Regularization-based DA \textbf{(GR-DA)}} 
\label{Geometric alignment-based DA}
\textbf{GR-DA} methods provide a distinct geometric perspective to effectively reduce domain shifts by integrating the underlying cross-domain manifold structure and minimizing \textbf{Term.3} in Equation (\ref{eq:bound}). These methods can be distinguished based on their usage of graph embedding techniques or their adoption of subspace alignment approaches.

By employing graph embedding (\textbf{GE}) techniques, a series of prior \textbf{GE-DA} approaches explicitly align hidden manifolds across distinct feature spaces \cite{jing2020adaptive,zhang2019manifold} within a purpose-designed knowledge graph for synchronized manifold representation in \textbf{DA}. For instance, \textbf{ARTL} \cite{long2013adaptation} utilizes the \textbf{GE} strategy to align diverse feature spaces and optimize the \textbf{MMD} distance within a unified framework. To counteract the geometric structure distortion present in the original space, \textbf{MEDA} \cite{wang2018visual} focuses on learning the Grassmann manifold to enhance manifold representations. \textbf{ACE} \cite{jing2020adaptive} contends that previous research often overlooks discriminative effectiveness and proposes enhancing the graph model's discriminativeness by increasing within-class distances and reducing between-class distances across different sub-domains. Different from \textit{feature spaces alignment enforced} \textbf{GE-DA} approaches, another prominent \textbf{GE-DA} research branch explores the more effective utilization of the discriminative label space. This approach additionally reduces \textbf{Term.1} in Equation (\ref{eq:bound}) and leads to label space embedding enforced \textbf{GE-DA}. In optimizing the label space, \textbf{EDA} \cite{zhang2016robust} and \textbf{DTLC} \cite{li2020discriminative} enforce manifold structure alignment across both the feature space and the label space, resulting in discriminative conditional distribution adaptation. By incorporating label smoothness consistency (\textbf{LSmC}), \textbf{UDA-LSC} \cite{DBLP:journals/tip/HouTYW16} effectively captures the geometric structures of the underlying data manifold in the label space of the target domain. \textbf{DGA-DA} \cite{luo2020discriminative} further refines \textbf{UDA-LSC} by simultaneously addressing manifold regularization in the source domain. Combining the strengths of both \textbf{GE-DA} research paradigms, Luo \textit{et al.} \cite{luo2022attention} introduce a novel \textbf{ARG-DA}  method that unifies different feature and label spaces using a dynamic optimization approach for comprehensive manifold unification-based \textbf{DA}.

In addition to \textit{graph embedding} based \textbf{DA} methods, an increasing number of \textbf{DA} techniques, such as those highlighted in \cite{DBLP:journals/corr/LuoWHC17,DBLP:journals/ijcv/ShaoKF14,DBLP:journals/tip/XuFWLZ16,sun2016return,DBLP:journals/tnn/DingF18,DBLP:conf/iccv/FernandoHST13}, place a strong emphasis on aligning the underlying data subspace instead of the graph model between the source and target domains for effective \textbf{DA}. These methods introduce low-rank and sparse constraints into \textbf{DA}, aiming to extract a low-dimensional feature representation where target samples can be sparsely reconstructed from source samples \cite{DBLP:journals/ijcv/ShaoKF14} or interleaved with source samples \cite{DBLP:journals/tip/XuFWLZ16}, thereby aligning the geometric structures of the underlying data manifolds. Some recent \textbf{DA} methods, such as \textbf{RSA-CDDA} \cite{DBLP:journals/corr/LuoWHC17} and \textbf{JGSA} \cite{Zhang_2017_CVPR}, further propose unified frameworks to reduce the shift between domains statistically and geometrically. \textbf{HCA} \cite{liu2019homologous} enhances \textbf{JGSA} by introducing a homologous constraint on the two transformations for the source and target domains, respectively, to establish a connection between the transformed domains and alleviate negative domain adaptation.

Despite the significant progress achieved by \textbf{GR-DA} methods in mitigating \textbf{Term.3} of Equation (\ref{eq:bound}), the sole alignment of data geometric structures across domains does not inherently ensure the theoretical effectiveness necessary to reduce the existed domain shift as stipulated in \textbf{Term.2} of Equation (\ref{eq:bound}). Furthermore, akin to \textbf{STA-DA} approaches, \textbf{GR-DA} methods also contend with considerable computational efficiency challenges when dealing with large datasets \cite{peng2019moment}, which restricts their practical applicability.

\subsection{Deep Domain Adaptation}
\label{subsect: Deep DA}

\textbf{DL-DA} techniques offer a promising solution to cross-domain tasks with a large number of samples. These methods utilize the \textit{'batch learning'} approach to dynamically sample the entire cross-domain dataset and optimize the overall function through an \textit{end-to-end} training strategy, resulting in discriminative feature representations and improved \textbf{DA} performance. Generally, traditional \textbf{DL-DA} techniques are typically categorized into two main research paradigms based on whether they incorporate an adversarial learning mechanism. In contrast, some recent \textbf{DA} methods expand upon generative models to approach Markov process-informed optimization for effective \textbf{DA}.

\subsubsection{Statistic Matching-based DA \textbf{(DL-STA-DA)}}
\label{Statistic matching-based DA}

 {These approaches share similar spirits  with \textbf{STA-DA} in reducing domain shift through various statistic measurements. However, \textbf{DL-STA-DA} advances this by incorporating the \textbf{DL} paradigm, moving beyond the shallow functional learning of \textbf{STA-DA}. This incorporation of popular deep frameworks, such as ResNet \cite{He2015} and AlexNet \cite{krizhevsky2012imagenet}, enables highly discriminative feature representations, leading to significantly improved performance.}

 {For instance, the well-known \textbf{DAN} \cite{long2015learning} reduces marginal distribution divergence by incorporating a multi-kernel MMD loss in the fully connected layers of AlexNet. \textbf{JAN} \cite{DBLP:conf/icml/LongZ0J17} builds on \textbf{DAN} by jointly minimizing the divergence of both marginal and conditional distributions. \textbf{D-CORAL} \cite{sun2016deep} takes this further by introducing second-order statistics into the AlexNet \cite{krizhevsky2012imagenet} framework, enhancing the effectiveness of the \textbf{DA} strategy. In contrast to previous research, \textbf{FDA} \cite{yang2020fda} and \textbf{MSDA} \cite{he2021multi} offer novel insights into reducing domain shift. \textbf{FDA} suggests that dissimilarities across domains can be measured using the low-frequency spectrum, while \textbf{MSDA} focuses on aligning cross-domains by reducing the \textbf{LAB} feature space representation. Both approaches, by swapping contents of different low-frequency spectra \cite{yang2020fda} or \textbf{LAB} feature space representations \cite{he2021multi} among cross-domains, significantly reduce domain shift. Additionally, \textbf{LMMD} \cite{zhu2020deep} introduces a novel local \textbf{MMD} measurement to effectively address sub-domain adaptation for conditional alignment in \textbf{DA}. \textbf{Mire} \cite{DBLP:conf/nips/ChenT0Z0Y22} defines a semantic anchor as the centroid of category-wise features within each domain, with the capability to reason over the global semantic topology for semantic knowledge-aware \textbf{DA}. Finally, \textbf{DRDA} \cite{DBLP:journals/tip/HuangWCZZ23} specifically designs global anchors to achieve cross-domain marginal distribution alignment, with a focus on both statistical and geometric alignment aspects.}

The fundamental principle of \textbf{DL-STA-DA} is to reduce domain divergence within kernel feature spaces \cite{belkin2018understand} through statistical measurements. The efficacy of this approach relies heavily on the thoughtful design of these statistical measurements. Given that \textbf{DL-STA-DA} methods are sensitive to the choice of statistical measurements, there has been a rising interest in adversarial learning-based \textbf{DA} techniques. These methods offer an automated optimization approach, eliminating the need for explicit calculation of prior distribution for likelihood maximization, thus moving away from manual, handcrafted approaches.

\subsubsection{Adversarial Loss-based DA}
\label{Adversarial loss-based DA}
These methods adopt the merits of adversarial learning \cite{goodfellow2014generative} to fool the deep model in recognizing the source and target domain samples, thus leading the cross-domain sample features indistinguishable \textit{w.r.t} the domain labels through an adversarial loss on a domain classifier \cite{ganin2016domain,tzeng2017adversarial,pei2018multi}. \textbf{DANN} \cite{ganin2016domain} and \textbf{ADDA} \cite{tzeng2017adversarial} learn a domain-invariant feature subspace by reducing the marginal distribution divergence. \textbf{MADA} \cite{pei2018multi} additionally makes use of multiple domain discriminators, thereby aligning conditional data distributions.  In contrast, \textbf{DSN} \cite{bousmalis2016domain} achieves domain-invariant representations by explicitly separating the similarities and dissimilarities in the source and target domains. \textbf{GVB} \cite{cui2020gradually} introduces a novel gradually vanishing bridge method to improve the feature representation on the generator and discriminator concurrently. \textbf{MADAN} \cite{zhao2019multi} explores knowledge from different multi-source domains to fulfill \textbf{DA} tasks. \textbf{CyCADA} \cite{pmlr-v80-hoffman18a} addresses the distribution divergence using a bi-directional \textbf{GAN}-based training framework.

The intriguing principle of Adversarial Loss-based \textbf{DA} approaches is intuitive and easy to implement, which has accelerated the rapid development of this research paradigm. However, these models are known for their potentially unstable training due to their adversarial nature, which lacks sufficient guidance \cite{bond2021deep, yang2022survey} for explicit likelihood distribution optimization.

\subsubsection{Generative Model-based DA}
\label{Generative Model-based DA}

In contrast to the discriminative models (${f_d}({\bf{x}},\theta ) \mapsto p({\bf{c}})$) commonly used in solving \textbf{DA} tasks, recently, the newly developed generative model (${f_g}(\theta ) \mapsto p({\bf{x}},{\bf{c}})$) in \textbf{DA} has also gained prominence in the field of \textbf{DA}, present an alternative approach. The rationale behind this is that its regularized functional learning \cite{lasserre2006principled, DBLP:conf/ijcai/JiangZTTZ17} not only approximates the label knowledge '${\bf{c}}$' as discriminative models do but also reversely generates samples with high fidelity. This ensures the preservation of key information during dynamic model training and making it a more reliable approach for \textbf{DA}.

Following the principle of Markov process-informed generative model optimization, \textbf{DDA} \cite{gao2023back} adopts a diffusion model to harmoniously blend different domains by deteriorating the data structure within the final noisy spaces. Therefore, the reversely generated cross-domain samples are assumed to be projected to the identical data distribution. Later on, Kamil \textit{et al.} \cite{deja2023learning} elaborated on classifier guidance to further approach conditional distribution-aware \textbf{DA}. Using a probabilistic graphical model, Zhang \textit{et al.} \cite{zhang2020domain} consider \textbf{DA} as a problem of Bayesian inference in graphical models. \textbf{VDI} \cite{xu2023domainindexing} further enhances the variational Bayesian framework by incorporating domain indices from multi-domain data, offering fresh insights into domain relationships and enhancing \textbf{DA} performance.

Recently, generative model-based \textbf{DA} methods have gained significant attention. However, as a novel yet promising research direction, when compared to discriminative model-based \textbf{DA}, they still face challenges related to computational efficiency and the generation of redundant samples during the overall optimization process. More specifically, diffusion model-enforced \textbf{DA} \cite{gao2023back} strictly adheres to the Markov chain optimization, which requires substantial computational resources for the reverse functional transformation. Furthermore, Bayesian inference-optimized \textbf{DA} generally reduces the formulated evidence lower bound, resulting in the generation of redundant samples that are not useful for decision optimization as required in solving \textbf{DA} tasks.

\subsection{Discussion}
\label{Discussion}

Given the considerable challenges posed by the large quantity of cross-domain datasets in the development of \textbf{Shallow-DA}, our research primarily focuses on the \textbf{DL-DA} paradigm. The key innovation of \textbf{DL-DA} methods lies in their ability to simultaneously reduce the divergence of data distributions across domains and generate a discriminative feature representation of data within a unified end-to-end learning framework. This results in an optimized model that harnesses the benefits of \textbf{META} learning \cite{wei2021metaalign}, where leveraging different tasks strengthens the model and allows for the discovery of optimal hyperparameters. However, in comparison to the '\textbf{Shallow-DA}' approach, which emphasizes '\textit{global modeling}' \cite{belkin2003laplacian}, the '\textit{batch learning}' strategy \cite{goodfellow2016deep} can notably limit the overall understanding of global statistics and geometry, as discussed in \textbf{Issue 1} and \textbf{Issue 2} within cross-domain experimental scenarios. To address these issues, we introduce a novel \textbf{DL-DA} method called \textbf{G}lobal \textbf{A}wareness e\textbf{N}hanced \textbf{D}omain \textbf{A}daptation (\textbf{GAN-DA}). This method incorporates an orthonormalized prior feature distribution to foster global statistical and geometric awareness within the \textbf{DL-DA} optimization framework. \textbf{GAN-DA} seamlessly integrates the prior feature distribution with the batch learning mechanism, resulting in deep \textbf{DA} that is globally aware of statistics and geometry.

\section{The proposed method}
\label{The proposed method}

Sect.\ref{Notations and Problem Statement} introduces the notations and states the \textbf{DA} problem. Sect.\ref{subsection:Formulation} delineates our \textbf{GAN-DA} model and offers a comprehensive, step-by-step account of the methodology. Sect.\ref{Overall Optimization} elucidates the complete optimization process of the proposed \textbf{GAN-DA} model to achieve efficient cross-domain recognition.

\subsection{Notations and Problem Statement}
\label{Notations and Problem Statement}

Matrices are written as boldface uppercase letters. Vectors are written as boldface lowercase letters. For matrix ${\bf{M}} = ({m_{ij}})$, its $i$-th row is denoted by ${{\bf{m}}^i}$, and its $j$-th column by ${{\bf{m}}_j}$. A domain $D$ is defined as an \textit{l}-dimensional feature space $\chi$ and a marginal probability distribution $P(x)$, \textit{i.e.}, $\mathcal{D}=\{\chi,P(x)\}$ with $x\in \chi$.  Given a specific domain $D$, a  task $T$ is composed of a \textit{C}-cardinality label set $\mathcal{Y}$  and a classifier $f(x)$,\textit{ i.e.}, $T = \{\mathcal{Y},f(x)\}$, where $f({x}) = \mathcal{Q}( y |x)$ can be interpreted as the class conditional probability distribution for each input sample $x$. In unsupervised domain adaptation, we are given a source domain $\mathcal{D_S}=\{x_{i}^{s},y_{i}^{s}\}_{i=1}^{n_s}$ with $n_s$ labeled samples ${{\bf{X}}_{\cal S}} = [x_1^s...x_{{n_s}}^s]$, which are associated with their class labels ${{\bf{Y}}_S} = {\{ {y_1},...,{y_{{n_s}}}\} ^T} \in {{\bf{\mathbb{R}}}^{{n_s} \times C}}$, and an unlabeled target domain $\mathcal{D_T}=\{x_{j}^{t}\}_{j=1}^{n_t}$ with $n_t$  unlabeled samples ${{\bf{X}}_{\cal T}} = [x_1^t...x_{{n_t}}^t]$, whose labels  ${{\bf{Y}}_T} = {\{ {y_{{n_s} + 1}},...,{y_{{n_s} + {n_t}}}\} ^T} \in {{\bf{\mathbb{R}}}^{{n_t} \times C}}$ are unknown. Here, ${y_i} \in {{\bf{\mathbb{R}}}^c}(1 \le i \le {n_s} + {n_t})$ is a one-vs-all label hot vector in which $y_i^j = 1$ if ${x_i}$ belongs to the $j$-th class, and $0$ otherwise. We  define the data matrix ${\bf{X}} = [{{\bf{X}}_S},{{\bf{X}}_T}] \in {R^{l*n}}$ ($l$ = feature dimension; $n = {n_s} + {n_t}$ ) in packing both the source and target data. The source domain $\mathcal{D_S}$ and target domain $\mathcal{D_T}$ are assumed to be different, \textit{i.e.},  $\mathcal{\chi}_S=\mathcal{{\chi}_T}$, $\mathcal{Y_S}=\mathcal{Y_T}$, $\mathcal{P}(\mathcal{\chi_S}) \neq \mathcal{P}(\mathcal{\chi_T})$, $\mathcal{Q}(\mathcal{Y_S}|\mathcal{\chi_{S}}) \neq \mathcal{Q}(\mathcal{Y_T}|\mathcal{\chi_{T}})$. We also define the notion of \textit{sub-domain}, \textit{i.e.}, class,  denoted as ${\cal D}_{\cal S}^{(c)}$, representing the set of samples in ${{\cal D}_{\cal S}}$ with the class label $c$. It is worth noting that the definition of sub-domains in the target domain, namely ${\cal D}_{\cal T}^{(c)}$, requires a base classifier, to attribute  pseudo labels for the samples in ${{\cal D}_{\cal T}}$.

\textbf{MMD}: The  maximum mean discrepancy (\textbf{MMD})  is an effective non-parametric distance measure  that compares the distributions of two sets of data by mapping the data into Reproducing Kernel Hilbert Space\cite{borgwardt2006integrating} (\textbf{RKHS}). Given two distributions $\mathcal{U}$ and $\mathcal{V}$, the MMD between $\mathcal{U}$ and $\mathcal{V}$ is defined as:

	\begin{equation}\label{eq:MMD}
	\resizebox{0.7\hsize}{!}{$Dist({\mathcal U},{\mathcal V}) = {(\int_{\mathcal U} {\phi ({u_i})} {\rm{ }}{d_{{u_i}}} - \int_{\mathcal V} {\phi ({v_i})} {\rm{ }}{d_{{v_i}}})_{H}}$}
\end{equation}

\noindent   {where} ${\mathcal U}=\{ u_1, \ldots, u_{n_1}, \ldots \}$ and ${\mathcal V} = \{ v_1, \ldots, v_{n_2}, \ldots \}$ are two random variable sets from distributions $\mathcal{U}$ and $\mathcal{V}$, respectively, and $\mathcal{H}$ is a universal \textbf{RKHS} with the reproducing kernel mapping $\phi$: $f(x) = \langle \phi(x), f \rangle$, $\phi: \mathcal{X} \to \mathcal{H}$. In real practice, the \textbf{MMD} is estimated on finite samples:

 {
	\begin{equation}\label{eq:MMD1}
	\resizebox{0.8\hsize}{!}{$Dist({\mathcal U},{\mathcal V}) = \parallel \frac{1}{n_1} \sum^{n_1}_{i=1} \phi(u_i) - \frac{1}{n_2} \sum^{n_2}_{i=1} \phi(v_i) \parallel_{{H}}$}
\end{equation}}

\noindent   {where} ${u_i}$ and ${v_i}$ are independent random samples drawn from the distributions $\mathcal{U}$ and $\mathcal{V}$ respectively.

 {In this paper, we highlight the issue of global ignorance in deep learning-based domain adaptation methods during batch learning-based optimization. To address this limitation, we introduce "Global Awareness Enhanced Domain Adaptation" (\textbf{GAN-DA}), a novel approach that endows these models with an understanding of global statistical and geometric characteristics. This enhancement allows \textbf{GAN-DA} to surpass the limitations of traditional batch-based deep domain adaptation models, leading to more comprehensive \textbf{DA}.}

\subsection{Formulation}
\label{subsection:Formulation}

According to the theoretical error bound formalized in Eq.(\ref{eq:bound}), our proposed \textbf{GAN-DA} approach (Fig.\ref{fig:3}.(a)) commences with Source Domain Structure Risk Minimization (\textbf{SRM}) \cite{vapnik1999nature} in Sect.\ref{Discriminative statistic alignment} to explicitly reduce the classification error (\textbf{Term.1} of Eq.(\ref{eq:bound})) on the source domain. However, the \textbf{SRM} assumption cannot be guaranteed on the target domain due to the existing domain shift. To tackle this issue, we introduce a novel mechanism in Sect.\ref{Attention Regularized Laplace Graph}\&\ref{Global Geometric Structure Aware Manifold Unification}, namely global statistical and geometric structure-aware cross-domain distribution alignment (Fig.\ref{fig:3}.(c)), to explicitly reduce the cross-domain divergence (\textbf{Term.2}\&\textbf{Term.3} of Eq.(\ref{eq:bound})). Furthermore, we introduce a multilinear conditioning adversarial learning strategy in Sect.\ref{Adversarial Learning Enhanced Domain Alignment} (Fig.\ref{fig:3}.(d)) to search for the unification of the optimized feature representation and the label space for smooth adaptation. These strategies harmoniously blend global statistical/geometric knowledge with model optimization for our final \textbf{GAN-DA} approach.

\subsubsection{Source Domain Structural Risk Mnimization}
\label{Discriminative statistic alignment}

  {To minimize \textbf{Term.1} of Eq.(\ref{eq:bound}) for the source domain's Structural Risk Minimization (\textbf{SRM}) predictor \cite{vapnik1999nature}, we have formulated the predictive model as follows:
  \begin{equation}\label{eq:srm}
	\begin{array}{l}
f = \arg {\min _{f \in {{ H}_{\mathcal K}}}}l(f(x),y) + {\mathcal R}(f)\\
s.t.\;\;\;f:{x_{s}} \mapsto {y_{s}}
\end{array}
\end{equation}
the first term on the right-hand side of Eq.(\ref{eq:srm}) represents the supervised loss on the source domain samples. This term uses  Cross-Entropy Loss to optimize a well-structured generator (Fig.1.(a)) for accurate label predictions on the source domain. ${H_{\mathcal K}}$ represents  the learned kernel space within the selected non-linear generator, while ${\mathcal R}(f)$ denotes the regularization term, utilizing a weight decay to reduce the complexity of the learned model.  }

In \textbf{DA} tasks, the learned predictor is expected to discover the best non-linear mapping, $f:{x_{s,t}} \mapsto {y_{s,t}}$, that can be applied to both the source and target domains. However, the existing cross-domain distribution divergence leads to an unreliable target domain mapping, $f:{x_t} \mapsto {y_t}$, which motivates our subsequent research aimed at reducing this divergence through global statistical and geometric distribution alignment (Sect.\ref{Attention Regularized Laplace Graph} to Sect.\ref{Adversarial Learning Enhanced Domain Alignment}).

\subsubsection{Global Statistics Aware Distribution Alignment}
\label{Attention Regularized Laplace Graph}

The \textit{domain divergence} between the source domain $\mathcal{D_S}$ and target domain $\mathcal{D_T}$ is mathematically formulated as the divergence of cross-domain marginal and conditional probability distributions \cite{long2013transfer}:

 {		\begin{equation}\label{eq:term2}
	\resizebox{0.65\hsize}{!}{$\left\{ \begin{gathered}
  {\mathbf{1}}{\mathbf{.}}{\text{ }}\mathcal{P}({x_s}) \ne \mathcal{P}(x_t),\mathcal{Q}(y_s|x_s) \ne \mathcal{Q}(y_t|x_t) \hfill \\
  {\mathbf{2}}{\mathbf{.}}{\text{ }}x_s = x_t, y_s = y_t \hfill \\ 
\end{gathered}  \right.$}
\end{equation}}

\noindent   {Eq}.(\ref{eq:term2}.2) states that the source domain ($\mathcal{D_S}$) and the target domain ($\mathcal{D_T}$) share the same feature space (${\cal X}$) and label space (${\cal Y}$), while Eq.(\ref{eq:term2}.1) indicates that the two domains differ in terms of their marginal and conditional probability distributions.

\begin{itemize}
	\item  \textbf{MMD measured distribution divergence:}
\end{itemize}

Due to its wide applicability \cite{pan2011domain,luo2022attention,DBLP:conf/icml/LongZ0J17,long2013transfer} and solid theoretical foundation \cite{gretton2007kernel,gretton2012kernel}, in this research, we employ the maximum mean discrepancy (\textbf{MMD}) measurement to evaluate the distances between the expectations of the source domain/sub-domain and target domain/sub-domain. This allows us to quantify the existing domain divergence in \textbf{RKHS}. To be specific, we define \textbf{1)} the empirical distance of the source and target domains as $Dist_{{\cal M}{\cal M}{\cal D}}^m$; and \textbf{2)} the conditional distance $Dist_{{\cal M}{\cal M}{\cal D}}^c$ as the sum of the empirical distances between sub-domains in ${{\cal D}_{\cal S}}$ and ${{\cal D}_{\cal T}}$ with the same label.

	 {
\begin{equation}\label{eq:3.1.2}
	\resizebox{0.85\hsize}{!}{$\left\{ \begin{gathered}
{\bf{1}}.{\rm{ }}Dist_{\cal MMD}^m = {(\int_{{{\cal D}_{\cal S}}} {\phi ({x_i})} {d_{{x_i}}} - \int_{{{\cal D}_{\cal T}}} {\phi ({x_j})} {d_{{x_j}}})_{H}}\\
{\bf{2}}.{\rm{ }}Dist_{\cal MMD}^c = \sum\limits_{c = 1...C} {} {(\int_{{\cal D}_{\cal S}^{(c)}} {\phi ({x_i})} {d_{{x_i}}} - \int_{{\cal D}_{\cal T}^{(c)}} {\phi ({x_j})} {d_{{x_j}}})_{H}}
\end{gathered} \right.$}
\end{equation}	}

\noindent   {where} $C$ is the number of classes, $\mathcal{D_S}^{(c)} = \{ {\bf{x}_i}:{\bf{x}_i} \in \mathcal{D_S} \wedge y({\bf{x}_i}) = c\} $ represents the ${c^{th}}$ sub-domain in the source domain,  $n_s^{(c)} = {\left\| {\mathcal{D_S}^{(c)}} \right\|_0}$ is  the number of samples in the ${c^{th}}$ {source} sub-domain. Similarly, $\mathcal{D_T}^{(c)}$ and $n_t^{(c)}$ are defined for the target domain but using pseudo-labels. Consequently, in Eq.(\ref{eq:3.1.2}), the divergence between cross-domain marginal distributions and that between conditional distributions are reduced when minimizing $Dist_{{\cal M}{\cal M}{\cal D}}^m$ and $Dist_{{\cal M}{\cal M}{\cal D}}^c$, respectively.

	\begin{itemize}
	\item  \textbf{Global statistics ignored distribution alignment:}
\end{itemize}

In practical applications, the ideal cross-domain distribution divergence, as defined in Eq.(\ref{eq:3.1.2}), is assessed using "batch sampling" strategies within the \textbf{DL-DA} models \cite{Tzeng_2017_CVPR, long2015learning, Yan_2017_CVPR}. Consequently, the cross-domain divergence is reformulated as follows:

 {
\begin{equation}\label{eq:b-ali}
	\resizebox{0.9\hsize}{!}{$\left\{ \begin{gathered}
{\bf{1}}.{\rm{ }}Dist_{\cal MMD}^{m - batch} = {(\int_{{{\cal D}_{\cal S}}}^{batch} {\phi ({x_i})} {d_{{x_i}}} - \int_{{{\cal D}_{\cal T}}}^{batch} {\phi ({x_j})} {d_{{x_j}}})_{H}}\\
{\bf{2}}.{\rm{ }}Dist_{\cal MMD}^{c - batch} = \sum\limits_{c = 1...C} {} {(\int_{{\cal D}_{\cal S}^{(c)}}^{batch} {\phi ({x_i})} {d_{{x_i}}} - \int_{{\cal D}_{\cal T}^{(c)}}^{batch} {\phi ({x_j})} {d_{{x_j}}})_{H}}
\end{gathered} \right.$}
\end{equation}	}

\noindent   {where} the samples are randomly sampled within a single batch across domains rather than being completely utilized as in Eq.(\ref{eq:3.1.2}). Therefore, as discussed in Fig.\ref{fig:1}.(a), Eq.(\ref{eq:b-ali}) may easily generate asymmetric sampling across domains. For example, the source batch contains '1', '2', and '3', while the target batch contains '2', '3', and '4'. In such circumstances, the source '1' and target '4' are poorly aligned with other conditioned samples, resulting in negative transfer. To mitigate negative transfer, recent research such as \textbf{CDAN} \cite{long2018conditional} and \textbf{LMMD} \cite{zhu2020deep} have further developed the effectiveness of label space-enforced semantic knowledge for conditional distribution-aware deep \textbf{DA}. However, as visualized in Fig.\ref{fig:1}.(b), these methods are still inadequate in capturing global statistics, leading to unavoidable contradictions among different statistical properties within the randomly sampled batches.

	\begin{itemize}
	\item  \textbf{VAE-inspired prior distribution designing:}
\end{itemize}

An interesting question that arises is whether it's possible to provide a trained model with a global statistical perspective, even when dealing with batch sampling experiments. To address this question, we can draw inspiration from the elegant variational autoencoder \cite{kingma2013auto} (\textbf{VAE}), which extends the generative capabilities of the earlier autoencoder \cite{bengio2013representation}.

	\begin{figure}[h!]
		\centering
		\includegraphics[width=1\linewidth]{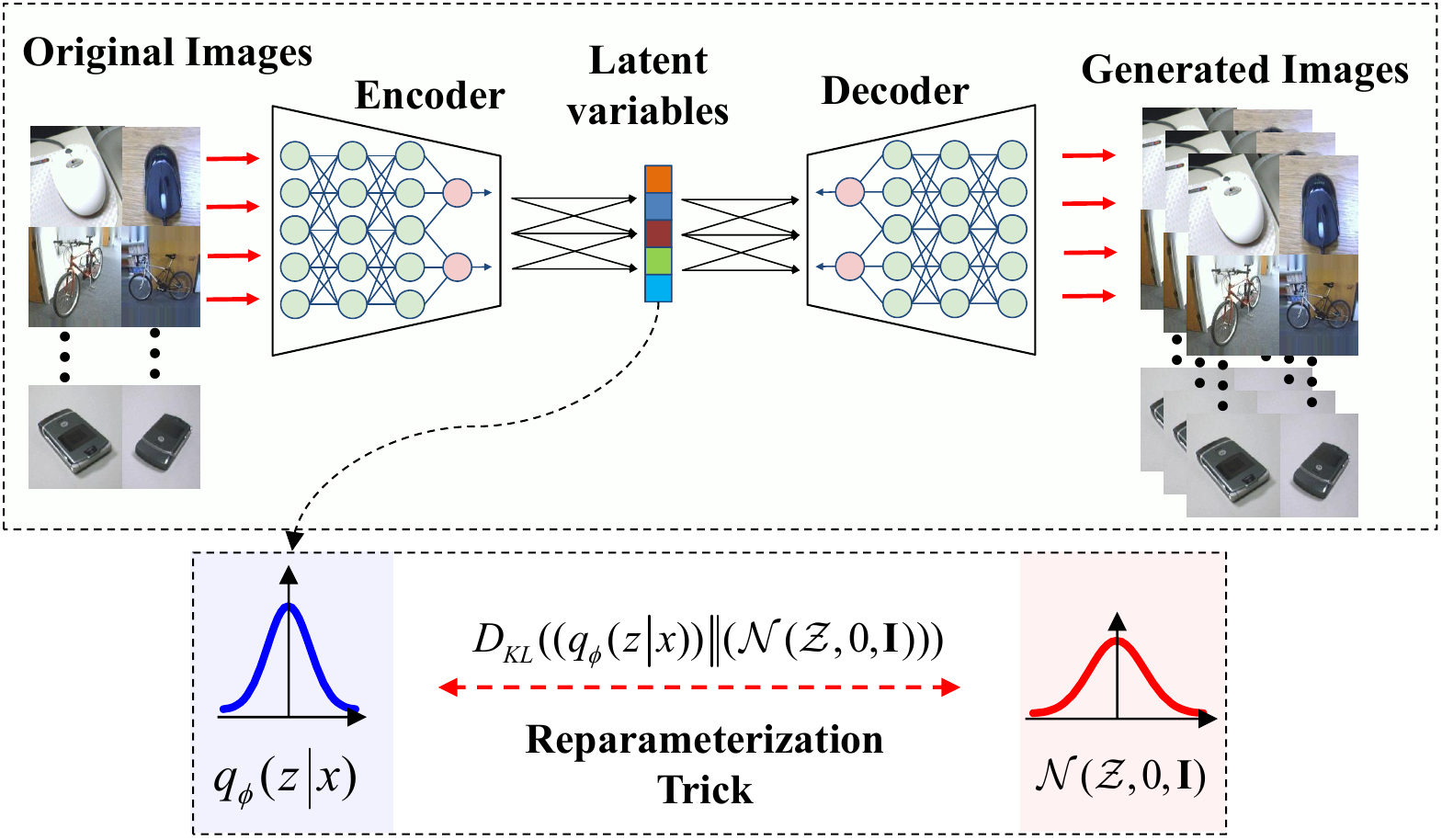}
		\caption {The learning procedure of the \textbf{VAE} involves approximating the distribution ${q_\phi }(z\left| x \right.)$ optimized over finite samples with a predefined data distribution ${p_\theta }(z) = {\mathcal{N}}(z,0,{\bf{I}})$ to enhance its generative capabilities.} 
		\label{fig:VAE}	
	\end{figure}

The vanilla variational autoencoder (\textbf{VAE}) proposes that the limited generation diversity of the autoencoder is a consequence of optimizing its latent feature representation, ${q_\phi}(z|x)$, using finite batch samples. This limitation results in insufficient semantic meaning in the latent feature representation, ${q_\phi}(z|x)$. To address this issue, \textbf{VAE} enforces ${q_\phi}(z|x)$ to align with a \textbf{predefined feature representation}, ${p_\theta}(z) = \mathcal{N}(z, 0, \mathbf{I})$, using the reparameterization trick to minimize the Kullback-Leibler (\textbf{KL}) divergence during model training. The predefined feature representation, $\mathcal{N}(z, 0, \mathbf{I})$, is supposed to cover the entire feature space, enabling the model to generate more diverse and realistic samples with infinite potential.

	\begin{itemize}
	\item  \textbf{Predefined feature representation motivated global statistics-aware distribution alignment:}
\end{itemize}

 {To address the global unawareness highlighted in Eq.(\ref{eq:b-ali}), inspired by \textbf{VAE}, we introduce a predefined feature representation (\textbf{PFR}) term, denoted as $\{ f_{pfr}^1...f_{pfr}^2...f_{pfr}^c\} $, as prior knowledge.} This \textbf{PFR} term proactively guides the mapping of cross-subdomain samples within different batches. The mathematical formulation of the mapping is as follows:

 {
\begin{equation}\label{eq:hf}
	\resizebox{0.7\hsize}{!}{$\begin{array}{l}
{\cal M}\{ {D_S},{D_T}\}  \mapsto \{ f_{pfr}^1...f_{pfr}^2...f_{pfr}^c\} \\
s.t.{({\cal M}\{ D_S^c,D_T^c\}  \mapsto p(f_{pfr}^c))_{(\forall c \in \{ 1...C\} )}}
\end{array}$}
\end{equation}	}

	\begin{figure}[h!]
		\centering
		\includegraphics[width=1\linewidth]{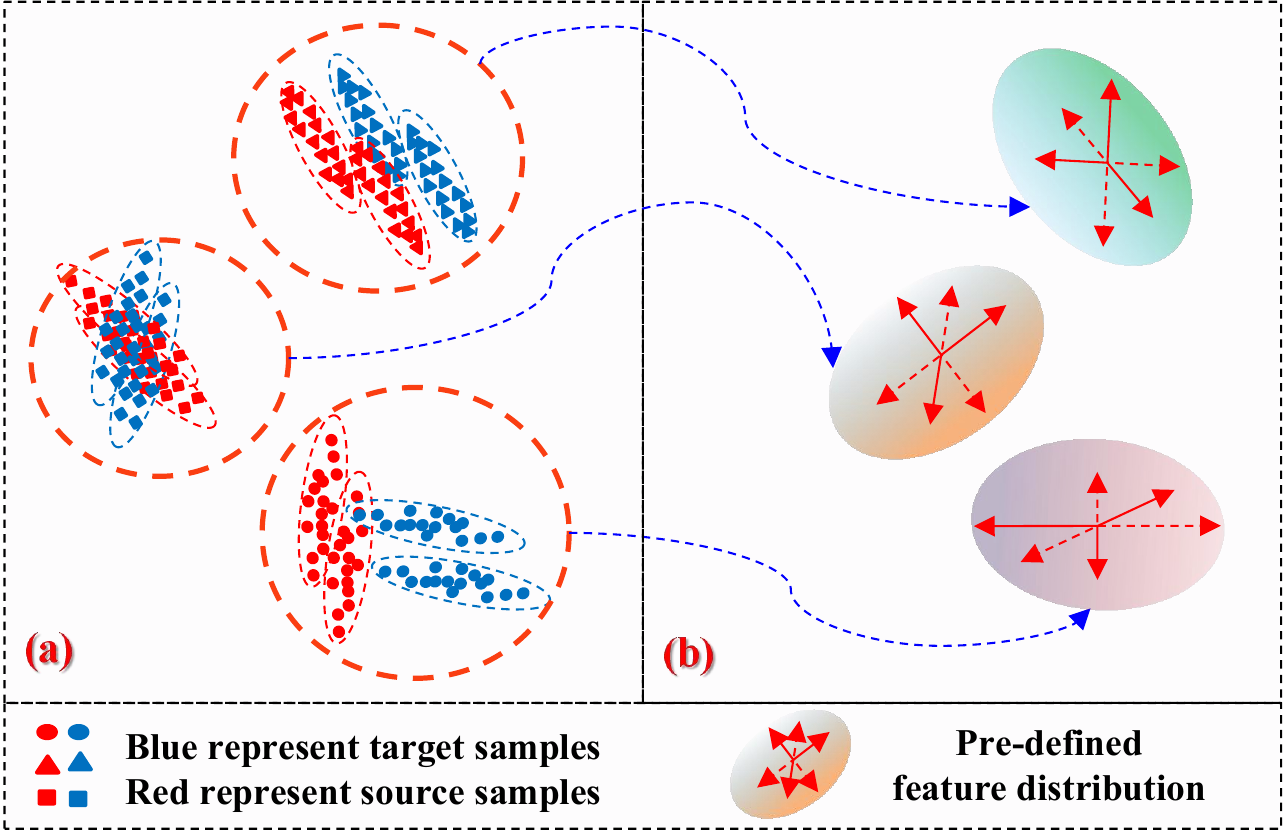}
		\caption {In Fig.\ref{fig:GSA} (a), we can observe that the cross sub-domain samples are collected from different batches. However, we synchronize these samples within the pre-defined feature distributions in Fig.\ref{fig:GSA} (b), which helps reduce the cross-domain divergence.} 
		\label{fig:GSA}	
	\end{figure}

 {The objective} of the optimized mapping ${\mathcal{M}}$ is to regress sub-domains that share similar labels, denoted as ${\{ {\cal D}_S^c,{\cal D}_T^c\} _{\forall c \in \{ 1...C\} }}$, to fixed positions in the predefined feature representation (\textbf{PFR}), represented as $p(f_{pfr}^c)_{\forall c \in \{ 1...C\}}$. This alignment of sub-domains with the \textbf{PFR} facilitates the alignment of conditional distributions across domains. As illustrated in Figure \ref{fig:GSA}, this synchronization of cross-subdomain samples from different batches through the \textbf{PFR} term significantly reduces cross-domain divergence and enhances global knowledge awareness. Consequently, using finite batched samples, the optimization problem in Eq.(\ref{eq:b-ali}) is reformulated as shown in Eq.(\ref{eq:hfmmd}):

 {
\begin{equation}\label{eq:hfmmd}
	\resizebox{0.8\hsize}{!}{$\left\{ \begin{gathered}
Dist_{G - {\cal M}{\cal M}{\cal D}}^{{\cal S} - batch} = \sum\limits_{c = 1...C} {} {(\int_{{\cal D}_{\cal S}^{(c)}}^{batch} {\phi ({x_i})} {d_{{x_i}}} - f_{pfr}^c)_H}\\
Dist_{G - {\cal M}{\cal M}{\cal D}}^{{\cal T} - batch} = \sum\limits_{c = 1...C} {} {(\int_{{\cal D}_T^{(c)}}^{batch} {\phi ({x_i})} {d_{{x_j}}} - f_{pfr}^c)_H}
\end{gathered} \right.$}
\end{equation}	}

\noindent   {where $f_{pfr}^c$ represents} the vectorized representation of the \textbf{PFR} term during actual optimization. \resizebox{0.18\hsize}{!}{$Dist_{G - {\cal M}{\cal M}{\cal D}}^{{\cal S} - batch}$} and \resizebox{0.18\hsize}{!}{$Dist_{G - {\cal M}{\cal M}{\cal D}}^{{\cal T} - batch}$} quantify the divergence between the batch-sampled source and target domains and the updated (\textit{vectorized}) \textbf{PFR} term. The minimization of Eq.(\ref{eq:hfmmd}) effectively reduces the \textit{marginal distribution divergence} across domains by aligning the cross-domain distributions with the \textbf{PFR} term. Furthermore, the \textit{conditional distribution divergence} is diminished by synchronizing the cross sub-domains with each vectorized sub-\textbf{PFR} term. Overall, Eq.(\ref{eq:hfmmd}) improves upon Eq.(\ref{eq:b-ali}) by explicitly introducing the \textbf{PFR} term, thereby reducing \textbf{Term.2} of Eq.(\ref{eq:bound}) while enhancing global statistical awareness.

\subsubsection{Global Manifold Unification Motivated Feature Representation Vectorization}
\label{Global Geometric Structure Aware Manifold Unification}

\begin{figure}[h!]
		\centering
		\includegraphics[width=1\linewidth]{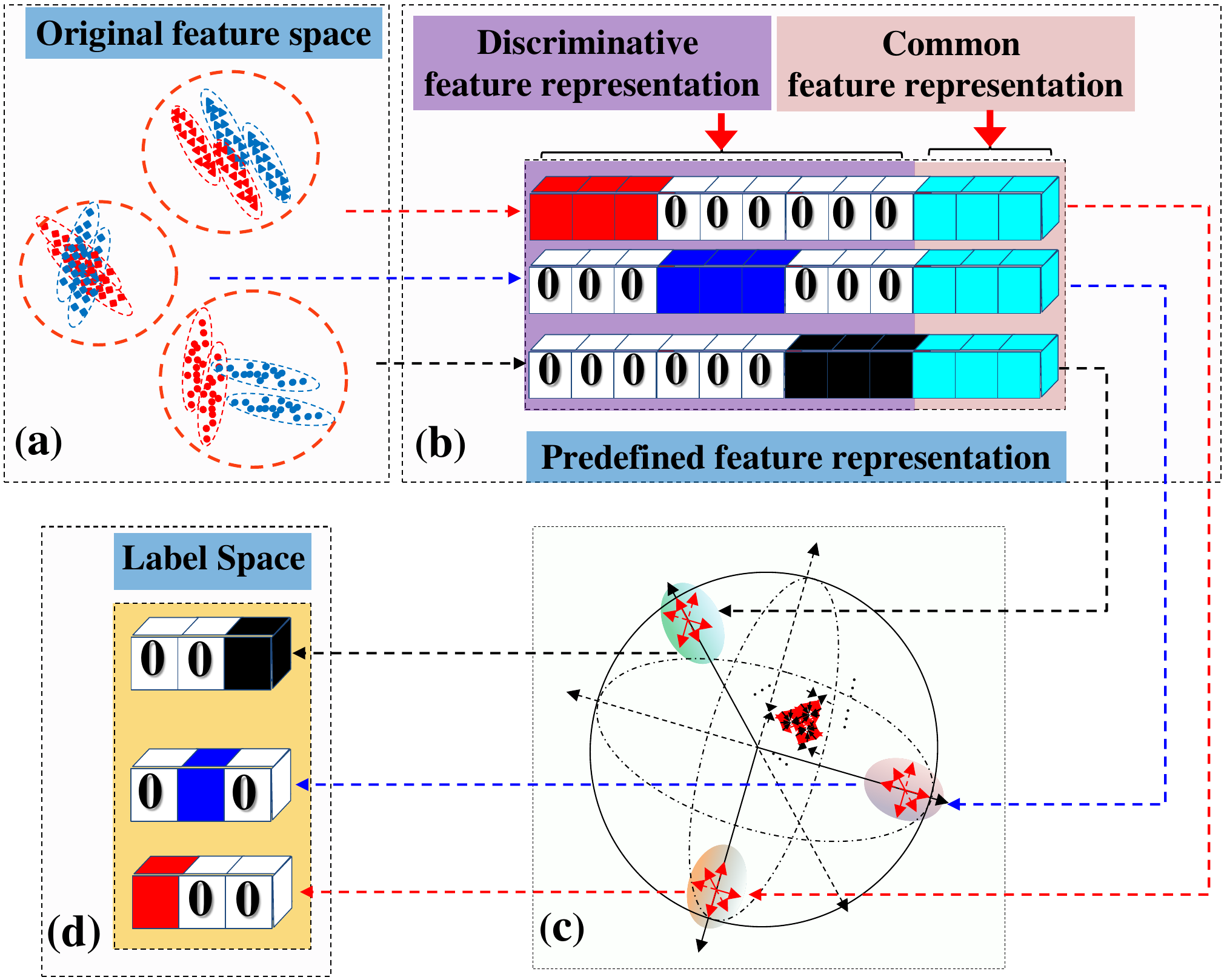}
		\caption {Fig.\ref{fig:GML} (a) shows cross sub-domain samples from different batches projected into the predefined feature space of Fig.\ref{fig:GML} (b), resulting in an orthogonal intermediate feature distribution (Fig.\ref{fig:GML} (c)). This transformation ensures that intermediate features are uncorrelated and can be interpreted independently. To enable proper cross-domain recognition, we use a proposed discriminative learning strategy to regress the orthogonal intermediate feature distribution into the one-hot label space (Fig.\ref{fig:GML} (d)). } 
		\label{fig:GML}	
	\end{figure}

To further enhance global manifold awareness, we posit that the designed \textbf{PFR} (Fig.\ref{fig:GML}.(b)) serves as a bridge connecting the \textit{original feature space} (Fig.\ref{fig:GML}.(a)) and the \textit{final label space} (Fig.\ref{fig:GML}.(d)), enabling \textit{comprehensive manifold unification} \cite{luo2022attention} and facilitating global manifold awareness. In this context, we assert that the \textbf{PFR} term exhibits the following three properties: \textbf{1)} Orthogonal property within the \textit{final label space}; \textbf{2)} Commonality preservation across sub-domains within the \textit{original feature space}; \textbf{3)} Discriminative functional learning for decision boundary optimization.

	\begin{itemize}
	\item  \textbf{Label Space Promoted Orthogonal Feature Representation:}
\end{itemize}

We propose that the designed \textbf{PFR} term (Fig.\ref{fig:GML}.(b)) serves as an intermediate feature space and should possess orthogonal properties, mimicking the one-hot label space and contributing to \textit{comprehensive manifold unification} across the original feature space to the final label space. This results in an updated mapping from Fig.\ref{fig:GML}.(b) to Fig.\ref{fig:GML}.(d), formulated as:

 {
\begin{equation}\label{eq:LD}
	\resizebox{0.62\hsize}{!}{$\begin{gathered}
{\cal M}'\{ ({f_{pfr}^c)_{\forall c \in \{ 1...C\} }}\}  \mapsto {({{\cal L}_c})_{\forall c \in \{ 1...C\} }}\\
s.t.\;\;\;\forall c,c' \in \{ 1...C\} ,c \ne c',{{\cal L}_c} \bot {{\cal L}_{c'}}
\end{gathered}$}
\end{equation}}

\noindent
 {
where ${{\cal L}_c}$ represents the $c$-th label subspace. As illustrated in Fig.\ref{fig:GML}.(d), the one-hot encoded ${{\cal L}_c}$ is typically represented as ${[0,0...\mathop 1\limits_c ...0,0]_{1*C}}$, and is orthogonal to ${{\cal L}_{c'}} = {[0,0...\mathop 1\limits_{c'} ...0,0]_{1*C}}$ when $c \ne c'$.  ${{\cal M}'}$ is the mapping required to optimize the non-linear relationship between $({f_{pfr}^c)_{\forall c \in \{ 1...C\}}}$ and ${({{\cal L}_c})_{\forall c \in \{ 1...C\} }}$. We argue that the \textbf{PFR} term should include a \textit{Vectorized Orthogonal Feature Representation} (\textbf{OFR}) term, denoted as ${F_{ofr}} = {(f_{ofr}^1...f_{ofr}^c...f_{ofr}^C)_{\forall c \in \{ 1...C\} }}$. This term specifically emulates the orthogonal characteristics of the label space, simplifying the optimization of ${\cal M}'$. In our study, we extend the dimension of the \textbf{OFR} term to $C*m$, where $m = \left\lfloor {D/(C + 1)} \right\rfloor$ (with '$\left\lfloor {} \right\rfloor $' representing the round down operation, and $D$ denoting the dimension of the original feature embedding from Fig.\ref{fig:3} (a)). This expansion accommodates the newly introduced \textbf{OFR} term (Fig.\ref{fig:GML}.(b)), enabling more extensive knowledge transfer from the original feature space. The term $C + 1$ represents  'C' distinct components corresponding to individual sub-domains, with the '1' indicating shared commonality across all C sub-domains. Consequently, the ${c^{th}}$ component of the new \textbf{OFR}, \textit{i.e.}, $f_{ofr}^c = {[(0...0)_{1*m}^{{1^{th}}}...(1...1)_{1*m}^{{c^{th}}}..(0...0)_{1*m}^{{C^{th}}}]_{1*(C*m)}}$, is responsible for propagating knowledge from the original ${c^{th}}$ sub-domain to the final label space using the ${c^{th}}$ \textbf{OFR} term. The vectorized \textbf{OFR} exhibits an orthogonal property, denoted as ${(f_{ofr}^c \bot f_{ofr}^{c'})_{\forall c,c' \in \{ 1...C\},c \ne c'}}$, which simplifies the regression to the one-hot label space.
}

However, the designed \textbf{OFR} term ignores the inherent similarities among different cross sub-domains within the original feature distribution. Therefore, incorporating these similarities \cite{bousmalis2016domain, yang2020fda} into the orthogonalized \textbf{OFR} term would be inappropriate and could result in negative transfer.

	\begin{itemize}
	\item  \textbf{Original Feature Space Induced Common Feature Representation:}
\end{itemize}

 { To address the issue of negative transfer, as depicted in Fig.\ref{fig:GML}.(b), we propose that the \textbf{PFR} should include a \textit{Vectorized Common Feature Representation} (\textbf{CFR}) term, denoted as ${F_{cfr}} = {(f_{cfr}^1...f_{cfr}^c...f_{cfr}^C)_{\forall c \in \{ 1...C\} }}$, which captures similar statistical and geometric properties across different sub-domains, \textit{i.e.}, ${(f_{cfr}^c \approx f_{cfr}^{c'})_{\forall c,c' \in \{ 1...C\} ,c \ne c'}}$. This term models the shared characteristics among sub-domains. Mathematically, the ${c^{th}}$ component of the \textbf{CFR} is defined as $f_{cfr}^c = {[1...1]_{1*n, n=D-(c*m)}}$, where $n$ represents the dimension of the \textbf{CFR} term. This allows for a $1*n$ dimensional feature space that encapsulates common knowledge across sub-domains. By integrating the \textbf{OFR} and \textbf{CFR} terms, Eq.(\ref{eq:hfmmd}) is reformulated as follows:}

 {
\begin{equation}\label{eq:hff}
	\resizebox{0.86\hsize}{!}{$\left\{ \begin{gathered}
Dist_{G - {\cal M}{\cal M}{\cal D}}^{{\cal S} - batch} = \sum\limits_{c = 1...C} {} {(\int_{{\cal D}_{\cal S}^{(c)}}^{batch} {\phi ({x_i})} {d_{{x_i}}} - f_{ofr}^c \oplus f_{cfr}^c)_{H}}\\
Dist_{G - {\cal M}{\cal M}{\cal D}}^{{\cal T} - batch} = \sum\limits_{c = 1...C} {} {(\int_{{\cal D}_T^{(c)}}^{batch} {\phi ({x_i})} {d_{{x_j}}} - f_{ofr}^c \oplus f_{cfr}^c)_{H}}
\end{gathered} \right.$}
\end{equation}	}

\noindent   {where} ${f_c}$ term in Eq.(\ref{eq:hfmmd}) is updated as \resizebox{0.18\hsize}{!}{$f_{ofr}^c \oplus f_{cfr}^c$}, the $ \oplus $ denotes the concatenation operation. In summary, \resizebox{0.5\hsize}{!}{${F_{pfr}} = {(f_{ofr}^c \oplus f_{cfr}^c)_{\forall c \in \{ 1...C\} }}$} represents the enhanced \textbf{PFR} term, contributing to achieving global statistical and geometric-aware distribution alignment.

\begin{itemize}
\item  \textbf{Discriminative Embedding Motivated Decision Boundary Optimization:}
\end{itemize}

When considering Eq.(\ref{eq:hff}), an intriguing question arises: \textit{'how do we quantify the importance of OFR and CFR terms in dynamically optimizing model training?'} To address this, we carefully review the entire training framework. During model training, the focus shifts from the \textbf{CFR} term, which captures similar elements within the original feature space, to the \textbf{OFR} term's role in mimicking the final orthogonal label space. To simulate this learning procedure, we introduce a parameter $\alpha = 1/(C + 1)$ to simulate dynamic importance weighting and enhance discriminative feature embedding. 
We assume that the '$C$' sub-domains contain '$C$' dissimilar components and only one similar component. Consequently, we scale \textbf{CFR} by $\alpha$ in each epoch to reduce similarities, while \textbf{OFR} is scaled by '$1/\alpha$' to emphasize discriminative traits and optimize the decision boundary. The reformulation of Eq.(\ref{eq:hff}) is presented in Eq.(\ref{eq:hffnew}), with the \textbf{PFR} term adjusted as \resizebox{0.85\hsize}{!}{${F_{pfr}} = ((1/{\alpha ^{epoch}})*f_{ofr}^c) \oplus ({\alpha ^{epoch}}*f_{cfr}^c){)_{\forall c \in \{ 1...C\} }}$}:

 {
            \begin{equation}\label{eq:hffnew}
	\resizebox{0.88\hsize}{!}{$\left\{ \begin{gathered}
Dist_{G - {\cal M}{\cal M}{\cal D}}^{{\cal S} - batch} = \hfill \\
\sum\limits_{c = 1...C} {} {(\int_{{\cal D}_{\cal S}^{(c)}}^{batch} {\phi ({x_i})} {d_{{x_i}}} - ((1/{\alpha ^{epoch}})*f_{ofr}^c) \oplus ({\alpha ^{epoch}}*f_{cfr}^c))_{H}}\\
Dist_{G - {\cal M}{\cal M}{\cal D}}^{{\cal T} - batch} = \hfill \\
\sum\limits_{c = 1...C} {} {(\int_{{\cal D}_T^{(c)}}^{batch} {\phi ({x_i})} {d_{{x_j}}} - ((1/{\alpha ^{epoch}})*f_{ofr}^c) \oplus ({\alpha ^{epoch}}*f_{cfr}^c))_{H}}
\end{gathered} \right.$}
\end{equation}	}

While the generated feature embedding aligns with the \textbf{PFR} term to improve global knowledge awareness, challenges arise from the divergence in dimensionality and orthogonality between the  ${\bf{PF}}{{\bf{R}}_{(1*D)}}$ and the final label spaces (${{\bf{Y}}_{(1*C)}}$). To tackle this challenge, we utilize adversarial learning to harmoniously merge these two spaces, promoting coherent functional learning.

\subsubsection{Adversarial Learning Enhanced Domain Alignment}
\label{Adversarial Learning Enhanced Domain Alignment}

 {Adversarial learning techniques \cite{Tzeng_2017_CVPR} offer a novel strategy for effectively merging feature representation ($f$) and conditional labels ($y$) to alleviate domain shift. Adhering to adversarial learning principles, successful cross-domain alignment occurs when the \textbf{discriminator} is unable to differentiate between generated features from the source or target domain. Indeed, in our \textbf{GAN-DA} framework, we use the multilinear conditioning operation mechanism introduced in the \textbf{CDAN} approach \cite{long2018conditional} to seamlessly unify the optimized feature representation $f$ with the inferred labels $y$ for joint distribution modeling. The optimization of the \textbf{generator} ($G$) and \textbf{discriminator} ($\mathcal{D}$) is mathematically expressed as follows:}

\begin{equation}\label{eq:adv}
	\resizebox{0.75\hsize}{!}{$\begin{gathered}
\mathop {\min }\limits_G \mathop {\max }\limits_D {L_{adv}} =  - \mathbb{E}[\sum\limits_{c = 1}^C {{\mathds{1}_{[{y_s} = c]}}\log \sigma (G({x_s}))} ]\\
 + \mathbb{E}[\log D({(f \otimes y)_s})] + \mathbb{E}[\log (1 - D({(f \otimes y)_t}))]
\end{gathered}$}
\end{equation}

\noindent   {where} the first term is dedicated to minimizing the structural risk of the source domain, with $\sigma$ corresponds to the softmax function, and $\mathds{1}$ serves as an indicator. The remaining two terms are aimed at optimizing the generator ($G$) and discriminator ($\mathcal{D}$) through an adversarial game, as described in Goodfellow \textit{et al.}'s work \cite{goodfellow2014generative}, to reach a model equilibrium. Notably, $\otimes$ represents the multilinear conditioning operation introduced by Long \textit{et al.} \cite{long2018conditional}. It leverages a kernel embedding technique to calculate the joint distribution of the optimized feature representations ($f$) and the inferred labels ($y$) using \textit{cross-covariance}, facilitating their harmonious blending. Overall, Eq. (\ref{eq:adv}) establishes a close relationship between feature representation and labels within a unified optimization framework, contributing to the seamless reduction of cross-domain divergence.

	\begin{figure}[h!]
		\centering
		\includegraphics[width=1\linewidth]{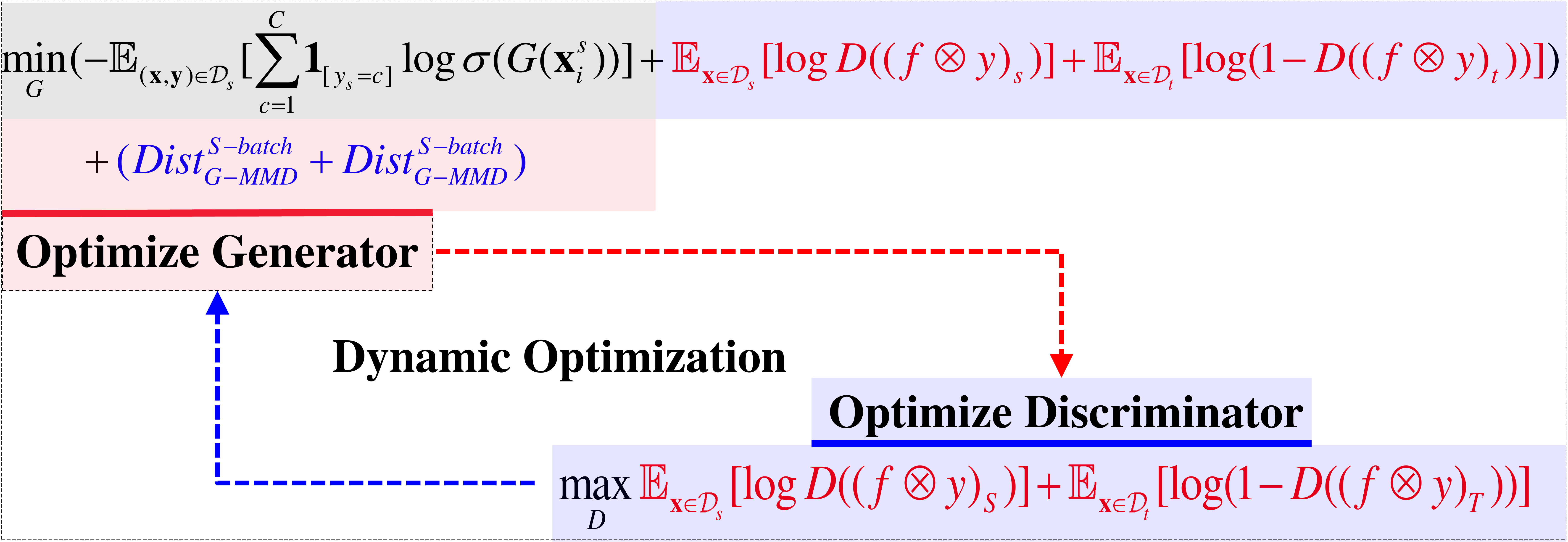}
		\caption { {The overall optimization procedure of the proposed \textbf{GAN-DA}} } 
		\label{fig:opti}	
	\end{figure}

\subsection{Overall Optimization}
\label{Overall Optimization}

 {As shown in Fig.\ref{fig:opti}, the overall optimization of the proposed \textbf{GAN-DA} framework involves optimizing both the generator ($\mathcal{G}$)  and discriminator ($\mathcal{D}$) to enhance global awareness in adversarial domain adaptation. The generator is optimized based on three distinct loss terms: \textit{Cross Entropy Loss}, \textit{Adversarial Loss}, and \textit{Global Statistics Alignment Loss}. These losses are mathematically formulated, distinguished by different color backgrounds, and visualized in the upper part of Fig.\ref{fig:opti}. \textit{Cross Entropy Loss} facilitates discriminative model training and directly reduces \textbf{Term.1} of the evidence lower bound in Eq.(\ref{eq:bound}), thus minimizing structural risk in the source domain. \textit{Adversarial Loss} contributes to \textbf{Term.2} of the evidence lower bound in Eq.(\ref{eq:bound}), explicitly reducing cross-domain distribution divergence. \textit{Global Statistics Alignment Loss} drives the generator to optimize \textbf{Term.3} by enforcing manifold unification across cross-domain decision boundaries, leveraging the predefined global feature geometric structure. The lower part of Fig.\ref{fig:opti} illustrates the optimization process within the discriminator, which further unifies cross-domain distributions. In summary, the overall optimization of \textbf{GAN-DA} integrates discriminative learning from labeled source domain samples, employs adversarial training to address domain shifts, and utilizes the specially designed \textbf{PFR} term to imbue the \textbf{DL-DA} model with comprehensive global knowledge awareness. This approach provides a novel global perspective, further enhancing \textbf{DA} performance.}

	\section{Experiments}
	\label{Experiments}

The experimental section provides a thorough evaluation of the proposed method, encompassing benchmarks and features (as outlined in Sect. \ref{subsection:Benchmarks and Features}), the experimental setup (as detailed in Sect. \ref{subsection: Experimental setup}), baseline methods (as listed in Sect.\ref{subsection:Baseline Methods}), and a discussion of the experimental results compared to the state of the art (as explored in Sect.\ref{subsection: Experimental Results and Discussion}). Additionally, the paper offers an analysis of the convergence and ablation study results of the proposed method, along with a visualization of the learned feature sub-spaces (as presented in Sect. \ref{Empirical Analysis}), providing a more profound insight into the proposed approach.

	\subsection{ Dateset Description}
	\label{subsection:Benchmarks and Features}

\textbf{Office-31}: The \textbf{Office-31} dataset is a standard benchmark for \textbf{DA}, comprising over 4000 images from three distinct domains: \textit{Amazon} (\textbf{A}, images obtained from online websites), \textit{Webcam} (\textbf{W}, low-resolution images captured by web cameras), and \textit{Dslr} (\textbf{D}, high-resolution images taken with digital SLR cameras). We assess our method across all six transfer tasks: \textit{A} $\rightarrow$ \textit{W} $\dots$ \textit{W} $\rightarrow$ \textit{D}.

	\textbf{Digits}: For the \textbf{Digits} domain adaptation tasks, we follow the lead of previous research \cite{long2018conditional,li2020maximum} by exploring three digit datasets: \textit{MNIST}, \textit{USPS}, and Street View House Numbers (\textit{SVHN}). \textit{MNIST} (\textbf{M}) comprises 60,000 training samples and 10,000 test samples. \textit{USPS} (\textbf{U}) includes 7,291 training samples and 2,007 test samples. \textit{SVHN} (\textbf{S}) encompasses over 600,000 labeled digits cropped from street view images.

	\textbf{Office-Home}: The \textbf{Office-Home} dataset is a recently introduced \textbf{DA} benchmark \cite{venkateswara2017deep} consisting of four domains, each with 65 categories. These domains include \textit{Art} (\textbf{Ar}), \textit{Clipart} (\textbf{Cl}), \textit{Product} (\textbf{Pr}), and \textit{Real-World} (\textbf{Rw}). From these four domains, we create 12 \textbf{DA} tasks, namely \textit{Ar} $\rightarrow$ \textit{Cl}, and so forth up to \textit{Pr} $\rightarrow$ \textit{Rw}.

	\textbf{ImageCLEF-DA\footnote{\url{http://imageclef.org/2014/adaptation}}}: The \textbf{ImageCLEF-DA} dataset, which is derived from 12 classes shared by four public datasets, assigns each dataset as a domain. These datasets include \textit{Caltech-256} (\textbf{C}), \textit{ImageNet ILSVRC 2012} (\textbf{I}), and \textit{Pascal VOC} 2012 (\textbf{P}). We consider three domain combinations (\textit{i.e.}, \textbf{C}, \textbf{I}, \textbf{P}), resulting in six cross-domain tasks, such as \textit{C} $\rightarrow$ \textit{I}, and so forth up to \textit{I} $\rightarrow$ \textit{P}.

	\subsection{Experimental Setup}
	\label{subsection: Experimental setup}

 {For feature generation in the \textbf{Digits} datasets, we utilized a variant of the LeNet model \cite{pmlr-v80-hoffman18a}, while the ResNet-50 architecture \cite{He2015} was employed for feature extraction in the image datasets. The domain discriminator was structured as \textit{FC-ReLU-DROP-FC-ReLU-DROP-FC-Sigmoid}, and the predictor was implemented using a standard softmax classifier. During optimization for digit recognition, mini-batch stochastic gradient descent (\textbf{SGD}) was used with a weight decay of $5*{10^{ - 4}}$, a momentum of 0.9, and a batch size of 64, following a learning rate similar to that of \textbf{CDAN} \cite{long2018conditional}. For object recognition in the \textbf{Image} datasets, we adopted similar hyper-parameter settings as in previous research \cite{li2020maximum}. The parameters '$\alpha $' and '$1/\alpha $' were applied during the initial two to three epochs to balance the contributions of the \textbf{CFR} and \textbf{OFR} terms. In this study, {\emph{accuracy}} on the test dataset, as defined by Eq.(14), was used as the performance metric, consistent with its widespread application in previous research, such as \cite{long2015learning, long2018conditional, DBLP:journals/corr/LuoWHC17, long2013transfer, DBLP:journals/tip/XuFWLZ16}, \textit{etc}.}

	\begin{equation}\label{eq:accuracy}
		\begin{array}{c}
			Accuracy = \frac{{\left| {x:x \in {D_T} \wedge \hat y(x) = y(x)} \right|}}{{\left| {x:x \in {D_T}} \right|}}
		\end{array}
	\end{equation}
	where ${\cal{D_T}}$ is the target domain treated as test data, ${\hat{y}(x)}$ is the predicted label, and ${y(x)}$ is the ground truth label for the test data  $x$.

	\subsection{Baseline Methods}
	\label{subsection:Baseline Methods}
	The proposed \textbf{GAN-DA} method is compared with \textbf{twenty-three} popular methods from recent literatures:

	\begin{itemize}

		\item \textbf{Domain adaptation methods}: 
		(1) \textbf{DAN}  \cite{long2015learning}.
		(2) \textbf{ADDA} \cite{tzeng2017adversarial};
		(3) \textbf{DRCN} \cite{ghifary2016deep};
		(4) \textbf{CoGAN} \cite{liu2016coupled};
		(5) \textbf{CDAN}  \cite{long2018conditional}.
		(6) \textbf{MCD} \cite{saito2018maximum};
		(7) \textbf{CAT} \cite{deng2019cluster};
		(8) \textbf{TPN} \cite{pan2019transferrable};
		(9) \textbf{BSP}  \cite{chen2019transferability}.
		(10) \textbf{GTA} \cite{sankaranarayanan2018generate};
		(11) \textbf{CyCADA} \cite{pmlr-v80-hoffman18a};
		(12) \textbf{UNIT} \cite{liu2017unsupervised};
		(13) \textbf{MSTN} \cite{xie2018learning};
		(14) \textbf{ATM} \cite{li2020maximum};
  		(15) \textbf{DDC} \cite{DBLP:journals/corr/TzengHZSD14};
		(16) \textbf{RTN} \cite{long2016unsupervised};
		(17) \textbf{JAN} \cite{long2016deep};
		(18) \textbf{D-CORAL} \cite{sun2016deep};
		(19) \textbf{MADA} \cite{pei2018multi};
		(20) \textbf{GSDA} \cite{hu2020unsupervised};
  		(21) \textbf{SWD} \cite{lee2019sliced};
  		(22) \textbf{TADA}   \cite{wang2019transferable};
  		(23) \textbf{BSP}   \cite{wang2019transferable};
  		(24) \textbf{DOLL-DA}   \cite{luo2024discriminative};
    
	\end{itemize}
	
Whenever possible, the reported performance scores of the \textbf{twenty-four} methods from the literature are directly collected from their original papers or previous research  \cite{long2018conditional, chen2019transferability, pei2018multi, li2020maximum, sankaranarayanan2018generate}. These scores are assumed to represent their \emph{best} performance.

	\subsection{Experimental Results and Discussion}
	\label{subsection: Experimental Results and Discussion}

	\subsubsection{\textbf{Experiments on the Digits DataSet}}
	\label{subsubsection:Experiments on the CMU PIE dataset} 

\begin{figure}[h!]
		\centering
		\includegraphics[width=0.7\linewidth]{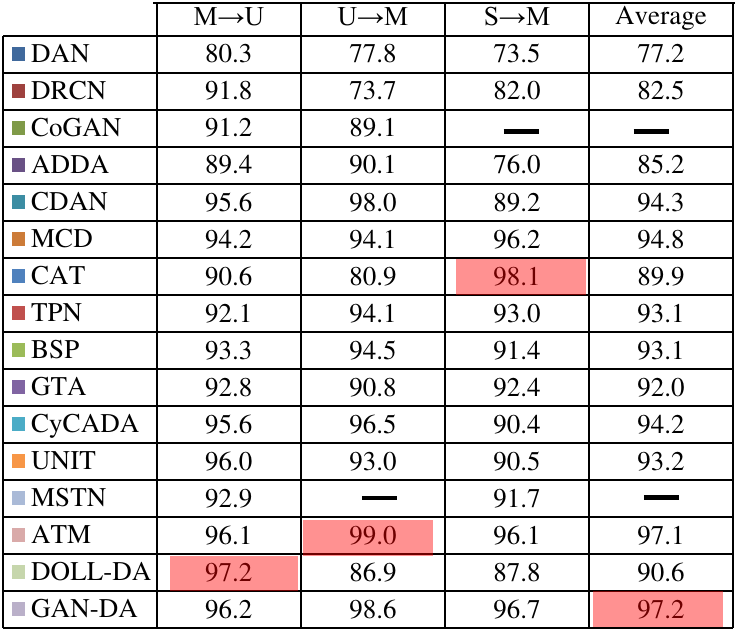}
		\caption {Accuracy${\rm{\% }}$ on Digital Image Datasets.} 
		\label{fig:Digits}
	\end{figure}

\textbf{MNIST}, \textbf{USPS}, and \textbf{SVHN} are three challenging handwriting datasets. While \textbf{MNIST} and \textbf{USPS} contain handwritten digits, \textbf{SVHN} is a collection of house numbers obtained from Google Street View images. Fig.\ref{fig:Digits} presents a synthesis of the experimental results for domain adaptation using these three datasets, highlighting the best results in red. Based on the results in Fig.\ref{fig:Digits}, the following observations can be made:

	\begin{itemize}
		
		\item The \textbf{SVHN-MNIST} dataset is a significant benchmark for domain adaptation, comprising 50,000 and 20,000 digit images from two distinct domains. Notably, the \textbf{CAT} \cite{deng2019cluster} approach leverages discriminative clustering structures in both domains to achieve better adaptation, resulting in the highest accuracy of $98.1\%$ for the \emph{S} $\rightarrow$ \emph{M} domain adaptation task. Our \textbf{GAN-DA} approach is tailored to implement global statistic/geometric aware domain adaptation for discriminative model training, which is a crucial component for achieving high performance. Importantly, our method achieved a noteworthy second-best performance of $96.7\%$.

		\item \textbf{ATM} \cite{li2020maximum} is an improvement over the popular \textbf{CDAN} method, featuring a specific design for maximum density divergence to enhance inter-domain divergence and intra-class density optimization. This approach achieved the best performance of $96.7\%$ for the \emph{U} $\rightarrow$ \emph{M} domain adaptation task and the second-best average performance across all proposed \textbf{DA} tasks. Similarly, the designed \textbf{GAN-DA} model optimizes both intra-class density and inter-domain divergence through a predefined feature embedding trick and discriminative learning mechanism, respectively. Furthermore, \textbf{GAN-DA} incorporates a globally enforced domain adaptation perspective, resulting in the highest average accuracy across all proposed \textbf{DA} tasks on the overall digits dataset.

	\end{itemize}

	\subsubsection{\textbf{Experiments on the Office-31 Dataset}}
	\label{subsubsection:Experiments on the Office-31 Dataset}
	
\begin{figure}[h!]
		\centering
		\includegraphics[width=1\linewidth]{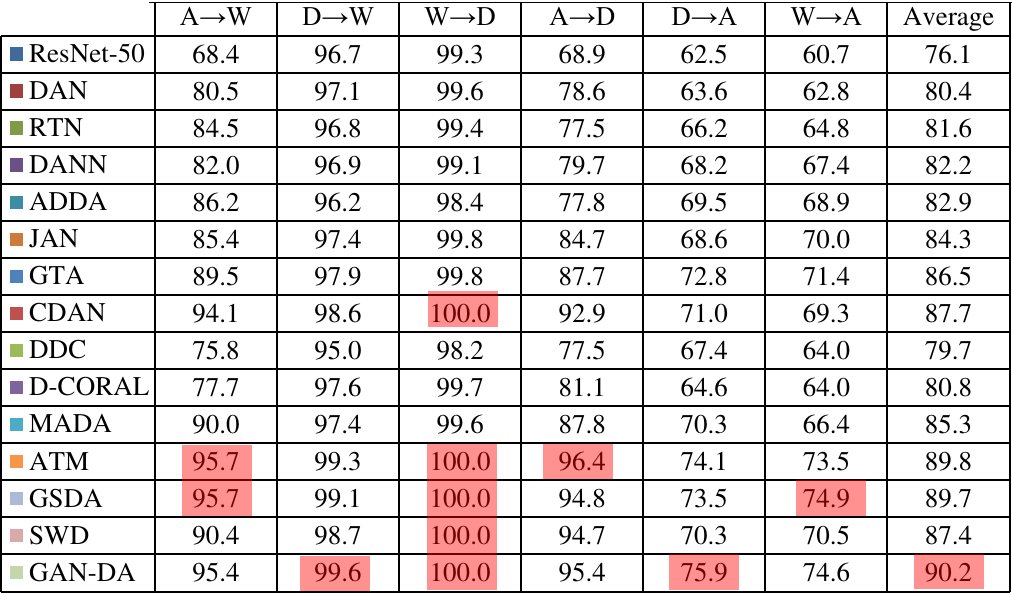}
		
		\caption {Accuracy${\rm{\% }}$ on Office-31 Dataset (ResNet-50).} 
		\label{fig:office31}
	\end{figure}

	\textbf{Office-31} is a highly popular dataset in the field of domain adaptation, featuring 31 objects commonly found in office settings such as monitors, keyboards, and printers. In Fig.\ref{fig:office31}, we present a comprehensive comparison of popular \textbf{DA} methods proposed for \textbf{Office-31} to evaluate their effectiveness in addressing domain shift. Specifically, \textbf{ResNet} serves as the baseline method for evaluating the effectiveness of the designed \textbf{DA} approaches in mitigating domain shift. Among these approaches, \textbf{ADDA} and \textbf{MADA} leverage adversarial learning mechanisms to minimize the divergence between cross-domain features and achieve feature alignment. \textbf{MADA} goes a step further by optimizing the decision boundary for multiple domain discriminators, enabling the model's sensitivity to the multimodal structures present in the data. The \textbf{GSDA} method, as presented in \cite{hu2020unsupervised}, reaches an accuracy of $89.7\%$ and secures the third position among all proposed \textbf{DA} methods. This success can be attributed to its comprehensive gradient synchronization technique, which randomly categorizes cross sub-domains into different groups, enabling the model to effectively leverage shared and unique features within the data, ultimately leading to improved adaptation performance. \textbf{ATM} achieves a commendable accuracy of $89.8\%$and secures the second position among the compared methods. This success is attributed to its specially designed maximum density divergence, which enhances both inter-domain divergence and intra-class density optimization. Our proposed \textbf{GAN-DA}, which takes into account global statistic/geometric properties for optimizing the \textbf{DA} model, outperforms all other methods by achieving the highest accuracy of $90.2\%$.

	\subsubsection{\textbf{Experiments on the ImageCLEF-DA Dataset}}
	\label{subsubsection:Experiments on the ImageCLEF-DA Dataset}
	
\begin{figure}[h!]
		\centering
		\includegraphics[width=1\linewidth]{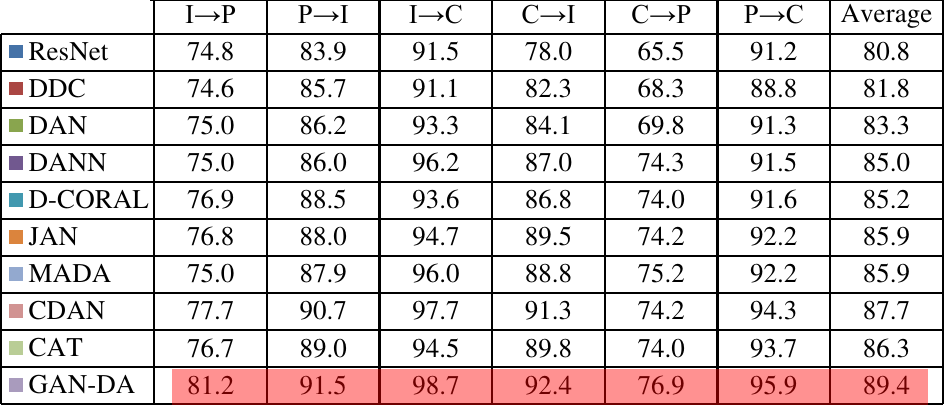}
		
		\caption {Accuracy${\rm{\% }}$ on ImageCLEF-DA Dataset (ResNet-50).} 
		\label{fig:clef}
	\end{figure}

\textbf{ImageCLEF-DA} contains three public datasets, \textit{i.e.}, Caltech-256 (\textbf{C}), ImageNet ILSVRC 2012 (\textbf{I}), and Pascal VOC 2012 (\textbf{P}), with each dataset treated as a separate domain. Each domain contains 600 images with 12 categories. The experimental results  in Fig.\ref{fig:clef}  follow the same experimental setup as in previous research \cite{long2018conditional, li2020maximum}. \textbf{DDC} utilizes the \textbf{MMD} measurement to explicitly reduce domain divergence and supervises structural risk in the source domain, achieving a certain level of accuracy. \textbf{DANN} builds upon this approach by incorporating adversarial learning strategies to attain domain-invariant features and achieves an accuracy of $85.0\%$. \textbf{CDAN} further addresses domain shift among the sub-domains and aligns the marginal distributions to improve accuracy, resulting in an accuracy of $87.7\%$. Our proposed method, \textbf{GAN-DA}, outperforms all other methods with an impressive accuracy of $89.4\%$, achieved by incorporating awareness of overall cross-domain statistic/geometric properties in the optimization process.

\subsubsection{\textbf{Experiments on the Office-Home Dataset}}
\label{subsubsection:Experiments on the Office-Home dataset}

 {\textbf{Office-Home}, introduced in \textbf{DAH} \cite{venkateswara2017deep}, presents a challenging benchmark for the \textbf{DA} task, comprising four domains with 65 object categories, resulting in 12 distinct \textbf{DA} tasks. In Fig.10,  the performance of the proposed \textbf{GAN-DA} is summarized alongside several state-of-the-art \textbf{DA} methods. \textbf{GSDA} achieves the second-best accuracy of $70.3\%$ by addressing the domain shift not only within the marginal/conditional distribution divergence but also the randomly categorized sub-domain groups. It introduces a novel hierarchical domain alignment strategy that addresses group-wise divergence for more comprehensive \textbf{DA}. The proposed \textbf{GAN-DA} achieves an accuracy of $70.6\%$, this performance can be attributed to the global perspective-enhanced statistic/geometric alignment and discriminative learning strategies for decision boundary optimization. Interestingly, the shallow \textbf{DA} method \textbf{DOLL-DA} shares a similar approach with \textbf{GAN-DA} and achieves superior accuracy. \textbf{DOLL-DA} explicitly regularizes the Maximum Mean Discrepancy (\textbf{MMD}) measurement for discriminative functional learning. Given the relatively small size of the \textit{Office-Home} dataset, \textbf{DOLL-DA} can leverage analytical matrix calculations for global optimization. Thus, for small datasets with discriminative deep features, the shallow \textbf{DA} method, such as \textbf{DOLL-DA}, slightly outperforms \textbf{GAN-DA}.}

	\begin{figure}[h!]
		\centering
		\includegraphics[width=1\linewidth]{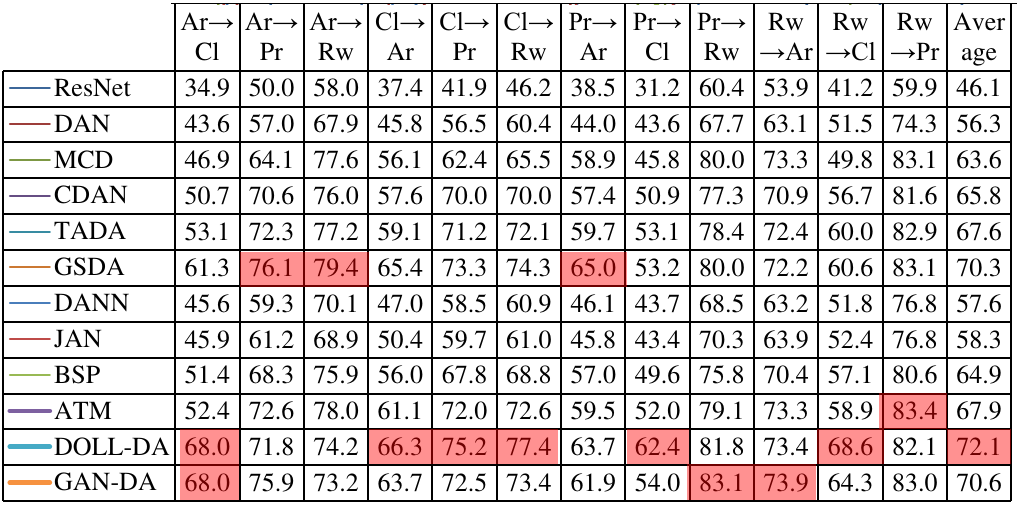}
		
		\caption { {Accuracy${\rm{\% }}$ on Office-Home Dataset (ResNet-50).} }
		\label{fig:HOME}
	\end{figure}

	\subsection{\textbf{Empirical Analysis}}
	\label{Empirical Analysis}

While the proposed \textbf{GAN-DA} achieves state-of-the-art performance across twenty-seven \textbf{DA} tasks encompassing four datasets, an intriguing question arises: how is the final decision boundary optimized through the model's global awareness in the presence of clean/noisy cross-domain datasets (see sect.\ref{Decision Boundary})? Furthermore, in Sect.\ref{Ablation Study}, we present a detailed ablation study to quantify \textbf{GAN-DA} and its derived models. This study includes both quantitative analysis of recognition accuracy and qualitative analysis through visualization of the results.

	\subsubsection{\textbf{Visulalization of the Decision Boundary Optimization}}
	\label{Decision Boundary}

\begin{figure}[h!]
		\centering
		\includegraphics[width=1\linewidth]{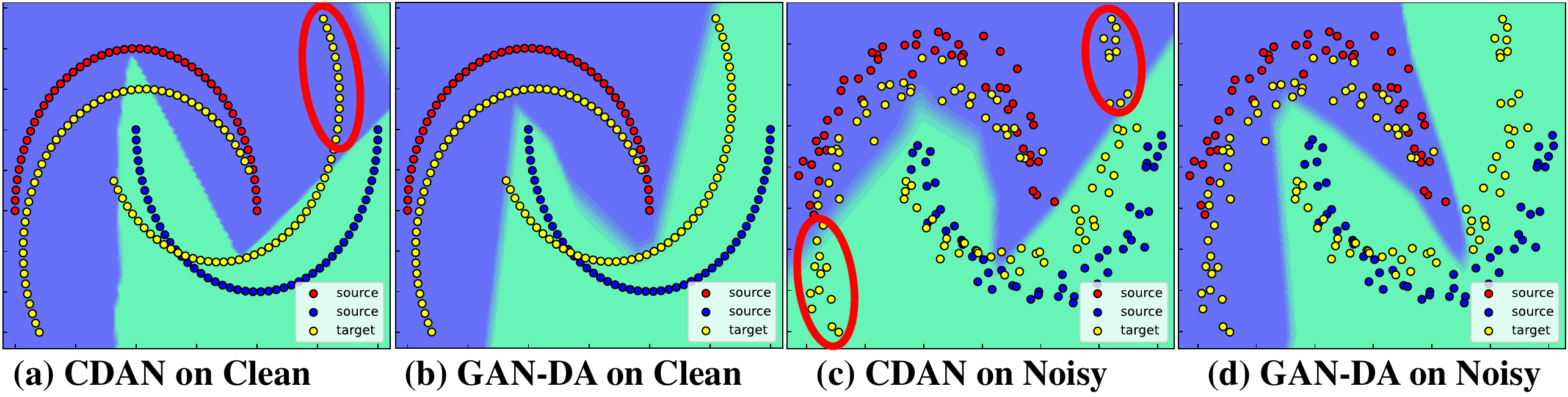}
		
		\caption {Figure.\ref{fig:moon}.(a\&b) and Figure.\ref{fig:moon}.(c\&d) provide visualizations of the performance of \textbf{CDAN} and \textbf{GAN-DA} on the double moon dataset under conditions of both small and large data variance, respectively.} 
		\label{fig:moon}
	\end{figure}

In Fig.\ref{fig:moon}, we present a comparison between the baseline method \textbf{CDAN} and our proposed \textbf{GAN-DA} in two experimental scenarios: a clean data distribution and a noisy data distribution. This visualization aims to demonstrate the effectiveness of our \textbf{GAN-DA} method in optimizing decision boundaries using the inter-twining moons 2D dataset. For the inter-twining moons 2D dataset, we set up the experiment as follows: The upper moon is labeled in red, and the lower moon is labeled in blue, with labels 0 and 1, respectively, representing the source domain. To create the target samples, we introduced a domain gap by rotating the source domain distribution by 35 degrees, resulting in 100 samples per class. The feature generator used in the experiment was a 4-layered fully-connected network, while the classifier and discriminator employed 3-layered fully-connected networks. The batch size during training was set to 8.

Upon analyzing Fig.\ref{fig:moon}.(a\&b), it becomes evident that our proposed \textbf{GAN-DA} method outperforms  \textbf{CDAN} by optimizing the decision boundary with appropriate margins, resulting in improved classification accuracy. This improvement can be attributed to the fact that our method takes into account the global geometric and statistical properties of the data, leading to a more robust and accurate decision boundary. Furthermore, as shown in Fig.\ref{fig:moon}.(c\&d), we observed that the noisy data distribution exhibits a more complex manifold structure. Interestingly, our proposed \textbf{GAN-DA} method still surpasses \textbf{CDAN} in this scenario. This observation indicates that the global geometric and statistical awareness of our method enables better decision boundary optimization across domains, even in the presence of more challenging and complex data distributions.

\begin{figure*}[h!]
	\centering
	\includegraphics[width=1\linewidth]{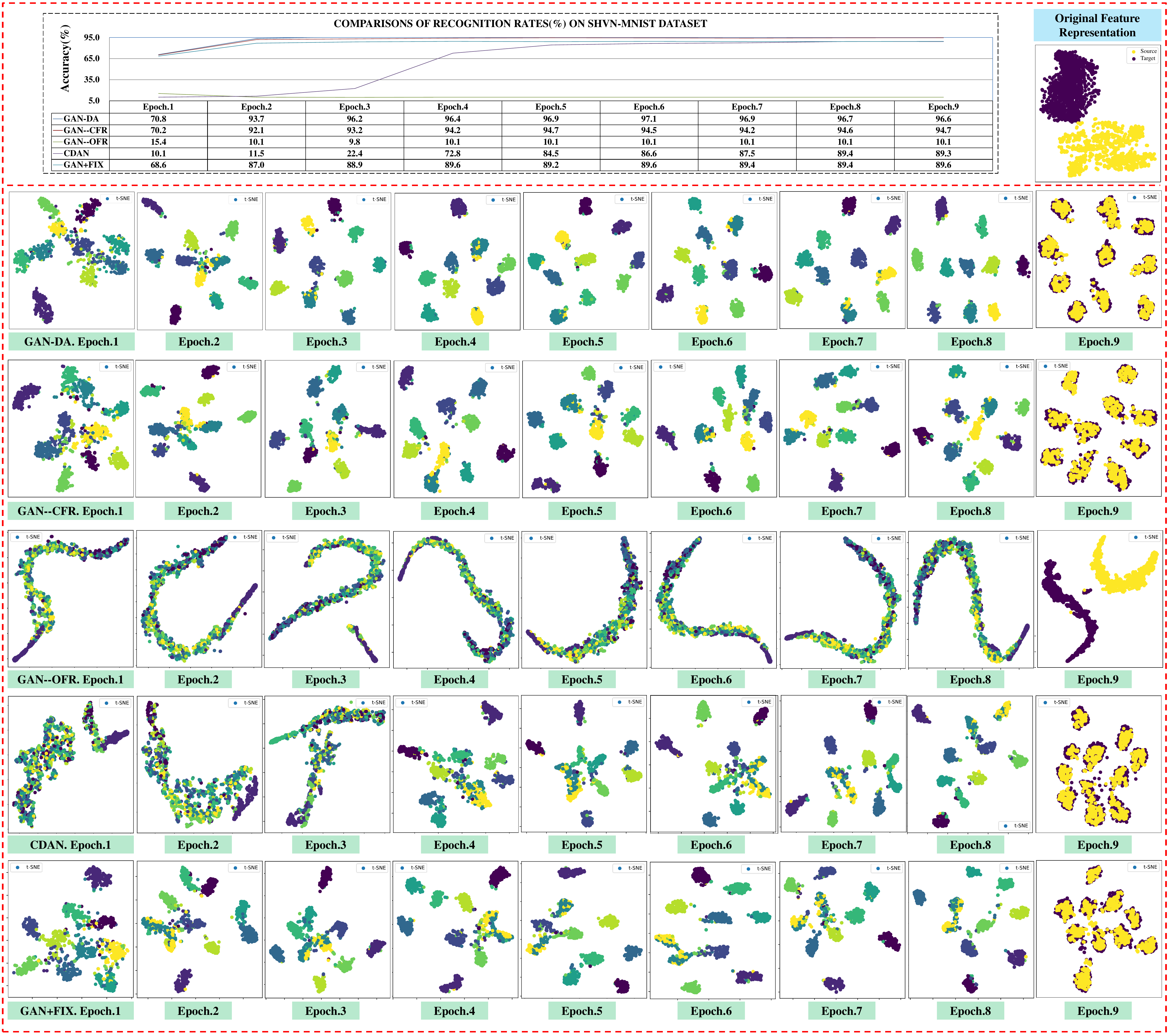}
	\caption {The upper table presents the accuracy of the five proposed \textbf{DA} models within the initial 9 epochs. The upper-left part of Fig.\ref{fig:abla} displays the original cross-domain divergence using the t-SNE feature visualization method. The lower section of Fig.\ref{fig:abla} exhibits the domain alignment results obtained by different \textbf{DA} methods in the final column labeled Epoch.9. To provide further insights into the characteristics of decision boundary optimization across various \textbf{DA} methods, we offer visualization results for the target domain spanning from the first to the eighth epoch.}
	\label{fig:abla}
\end{figure*}

	\subsubsection{\textbf{Ablation Study}}
	\label{Ablation Study}

While \textbf{GAN-DA} has demonstrated impressive performance compared to a variety of popular \textbf{DA} methods, its success can be attributed to its capacity to leverage global geometric and statistical awareness to align with the designed predefined feature representations. However, we are keen to explore the specific contributions of the \textbf{OFR}, and \textbf{CFR} components within the handcrafted feature. To gain a better understanding of the contributions of these key components in \textbf{GAN-DA}, we have proposed four derived models:

	\begin{itemize}
    	\item \textbf{CDAN}: This model eliminates the alignment of predefined feature representations, focusing on the assumed global statistical and geometric awareness.

		\item \textbf{GAN-CFR}: The \textbf{CFR} term in \textbf{GAN-DA} is set to a ${\bf{0}}$ matrix, removing the contribution of the \textbf{CFR} term and emphasizing the contribution of the \textbf{ORF} term. 
  
  		\item \textbf{GAN-ORF}: The \textbf{ORF} term in \textbf{GAN-DA} is set to a ${\bf{0}}$ matrix, highlighting the contribution of the \textbf{CFR} term. 
       
            \item \textbf{GAN+FIX}: In this model, the \textbf{CFR} term is fixed as a ${\bf{1}}$ matrix instead of gradually decreasing it, allowing for a better understanding of the shortcomings of the \textbf{GAN-ORF} model. 

  \end{itemize}

Overall, the utilization of these four derived models has facilitated a more profound comprehension of the significance of each fundamental component in the \textbf{GAN-DA} framework. Through a series of comprehensive experiments and the presentation of results in Fig. \ref{fig:abla}, encompassing both quantitative assessments of recognition accuracy and qualitative analyses via result visualizations, we can derive the following conclusions:

	\begin{itemize}

		\item Fig. \ref{fig:abla} presents the upper part table displaying the accuracy of the five proposed \textbf{DA} models during the first nine epochs. The upper-left section of Fig. \ref{fig:abla} showcases the original cross-domain divergence, as visualized using the t-SNE feature visualization method. The lower part of Fig. \ref{fig:abla} demonstrates the domain alignment achieved by different \textbf{DA} methods in the last column, denoted as "Epoch.9." To gain a better understanding of the decision boundary optimization characteristics across various \textbf{DA} methods, we offer visualization results for the target domain from the first to the eighth epoch. This approach places a greater emphasis on unsupervised target domain recognition, as it is deemed more crucial than supervised source domain classification.

    \item   {\textit{Predefined Feature Distribution Encourages Fast Convergence}: The models \textbf{GAN-DA}, \textbf{GAN-CFR}, and \textbf{GAN+FIX} demonstrated higher initial recognition accuracy (approximately 68\% to 70\%) during Epoch 1, compared to \textbf{CDAN}, which achieved only 10\% accuracy. Moreover, these models exhibited faster convergence rates, reaching convergence within 3 to 5 epochs, while \textbf{CDAN} required 8 epochs. This accelerated performance is attributed to the \textbf{PFR} term, which provides the model with essential prior knowledge, facilitating quicker global optimization.}

      \item  {\textit{Failure of ${\bf{0}}$ Space Projection in \textbf{GAN-OFR}:}  When the \textbf{OFR} term is initially set to \textbf{0} and the \textbf{CFR} term is reduced to a ${\bf{1}}*{10^{ - 3}}$ matrix after Epoch 3, the resulting feature embedding approaches an almost \textbf{0} space projection. This forced \textbf{0} space projection undermines the discriminative capabilities of the learned model, causing significant performance degradation. Specifically, the visualization results of \textbf{GAN-OFR} reveal that the excessive \textbf{0} space projection severely compromises the quality of the feature representation within the regularized functional learning.}

      \item   {\textit{Fixed CFR Term Approaches Marginal Distribution Alignment}: To address the failure of \textbf{GAN-OFR}, we introduced the \textbf{GAN+FIX} model, where the \textbf{CFR} term is fixed as a ${\bf{1}}$ matrix, similar to its application in \textbf{GAN-OFR}. This reformulation aims to unify cross-domain marginal distributions by maintaining a constant \textbf{CFR} term as the alignment objective. As a result, this approach significantly improves the accuracy of \textbf{GAN-OFR}, boosting it from 10.1\% to 89.6\%, and effectively prevents model collapse due to the \textbf{0} space embedding.}

    \item  {\textit{Global Knowledge-Aware DA Model}: The core idea behind the proposed \textbf{GAN-DA} is to enhance the \textbf{DL-DA} model with appropriate global statistical and geometric prior knowledge to minimize domain shift comprehensively. To achieve this, we specifically designed the \textbf{PFR} term, which contains the \textbf{OFR} and \textbf{CFR} components to reflect the characteristics of the final label space and avoid negative transfer, respectively. The ablation study demonstrates that the improved performance of \textbf{GAN-DA} over the \textbf{GAN-CFR}, \textbf{GAN-OFR}, and \textbf{CDAN} models highlights the contributions of the \textbf{OFR}, \textbf{CFR}, and their combination to the domain adaptation task.}
    
	\end{itemize}

	\section{Conclusion}
	\label{Conclusion}

 {In this research, we propose the \textbf{G}lobal \textbf{A}wareness e\textbf{N}hanced deep \textbf{D}omain \textbf{A}daptation (\textbf{GAN-DA}) method to address the issue of global knowledge unawareness in batch-learning-based \textbf{DL-DA} models. Our approach introduces a predefined feature representation (\textbf{PFR}) as an intermediate feature space, enabling the model to gain global statistical awareness. The \textbf{PFR} is further extended with \textbf{orthogonal feature representation} and \textbf{common feature representation} terms, promoting the unification of global manifold structures and optimizing decision boundaries for more discriminative \textbf{DA}. Our method outperforms existing domain adaptation techniques on standard benchmarks. We provide comprehensive analyses of \textbf{GAN-DA}, including examinations of decision boundaries, ablation studies, result visualizations, and convergence behavior, highlighting its effectiveness in overcoming global knowledge unawareness. Future work includes generalizing the introduced global knowledge mechanism to a variety of downstream tasks to enhance real-world computer vision applications such as detection, segmentation, and tracking.}

\ifCLASSOPTIONcaptionsoff
  \newpage
\fi

\tiny\bibliographystyle{plain}
\tiny\bibliography{cdda}

\begin{IEEEbiography}   [{\includegraphics[width=1in,height=1.25in,clip,keepaspectratio]{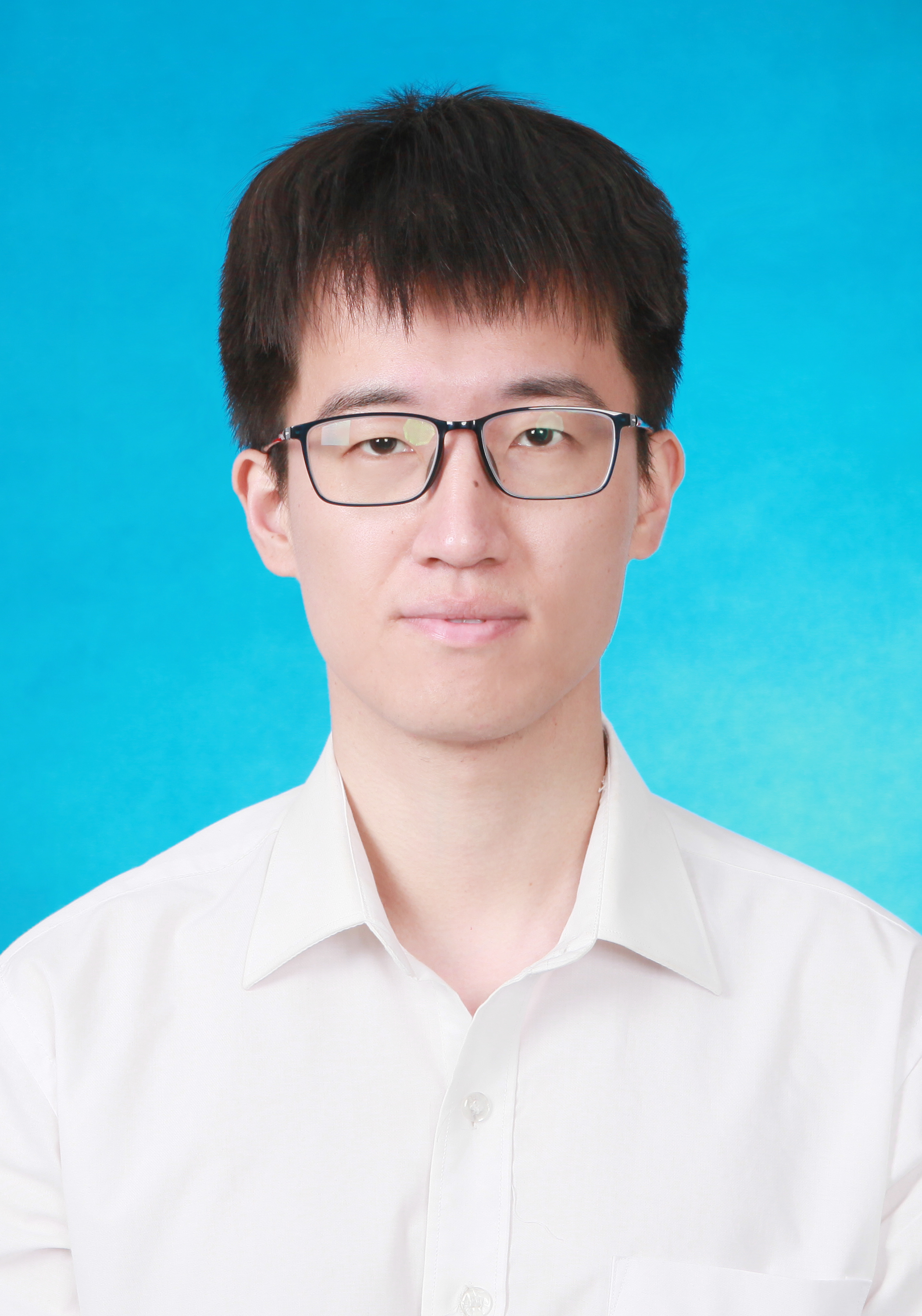}}]{Lingkun Luo} served as a research assistant and postdoc at Ecole Centrale de Lyon, Department of Mathematics and Computer Science, and was a member of the LIRIS laboratory. He is currently a research fellow at Shanghai Jiao Tong University. He has authored over 30 research articles, including publications in IJCV, ACM-CS, IEEE TIP, IEEE TCYB, IEEE TIFS and others. His research interests include machine learning, pattern recognition, and computer vision.
\end{IEEEbiography}

\begin{IEEEbiography}   [{\includegraphics[width=1in,height=1.25in,clip,keepaspectratio]{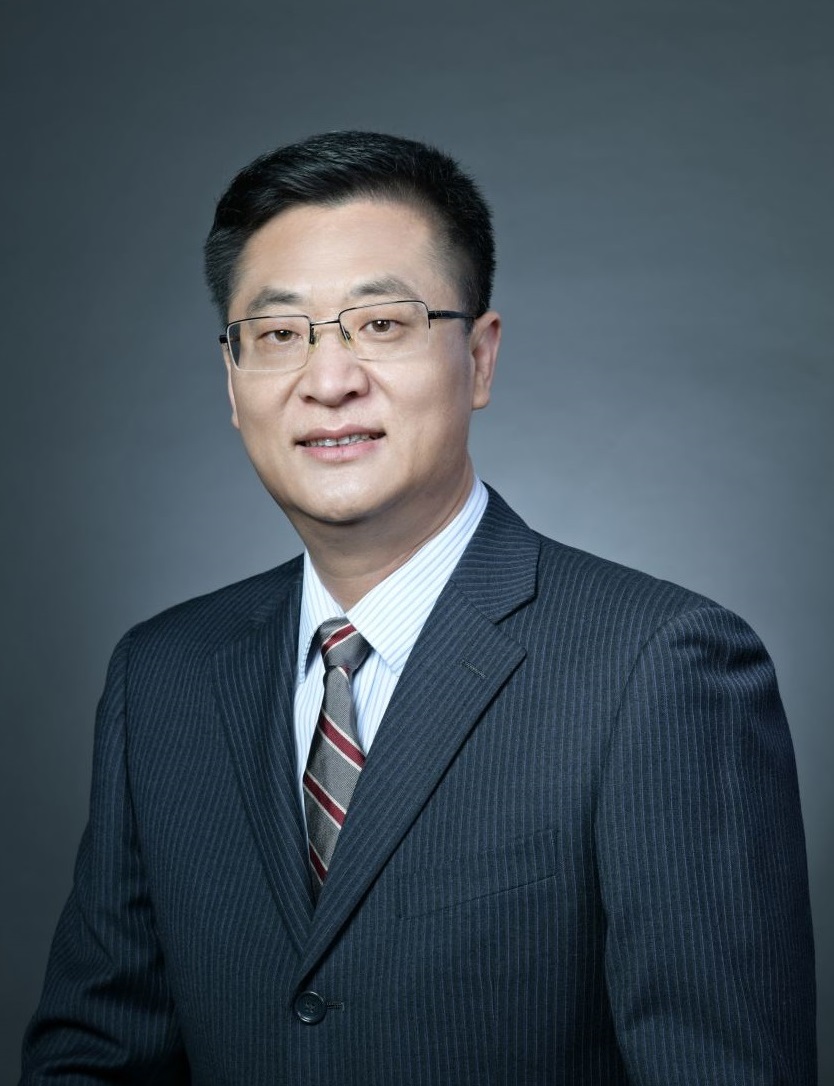}}] {Shiqiang Hu} earned his Ph.D. degree from Beijing Institute of Technology and currently serves as the dean of the School of Aeronautics and Astronautics at Shanghai Jiao Tong University. He has been at the helm of numerous research projects, including those funded by the National Science Foundation and the 863 National High Technology Plan. With over 300 publications to his name, he has also effectively mentored more than 20 Ph.D. students. Today, he holds the position of full professor at Shanghai Jiao Tong University. His research focus encompasses machine learning, image understanding, and nonlinear filters.
\end{IEEEbiography}

\begin{IEEEbiography}[{\includegraphics[width=1in,height=1.25in,clip,keepaspectratio]{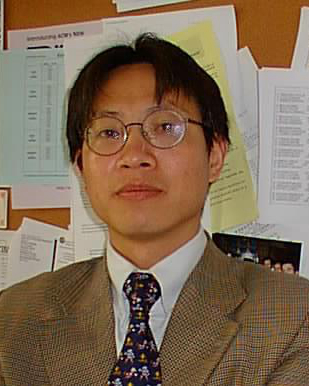}}]{Liming Chen} is a distinguished Professor at the Department of Mathematics and Computer Science, Ecole Centrale de Lyon, University of Lyon, France, and a senior member of the Institut Universitaire de France (IUF) through the chair of innovation since October 2022. He received his BSc in Mathematics and Computer Science from the University of Nantes in 1984, his MSc and PhD in computer science from the University Pierre and Marie Curie Paris 6 in 1986 and 1989 respectively. He was an associate professor at the Universit\'{e} de Technologie de Compi\`{e}gne before he joined \'Ecole Centrale de Lyon, \'Ecully as Professor in 1998. He served as the Chief Scientific Officer in the Paris-based company Avivias from 2001 to 2003, and the scientific multimedia expert in France Telecom R\&D China in 2005. He was the head of the Department of Mathematics and Computer science from 2007 through 2016. His current research interests include computer vision, machine learning, and multimedia with a particular focus on robot vision and learning since 2016. Liming has over 300 publications and successfully supervised over 40 PhD students. He has been a grant holder for a number of research grants from EU FP program, French research funding bodies and local government departments. Liming has so far guest-edited 5 journal special issues. He is an associate editor for Eurasip Journal on Image and Video Processing, area editor for Computer Vision and Image Understanding and a senior IEEE member.
\end{IEEEbiography}

\end{CJK*}
\end{document}